\documentclass{article}

\usepackage{arxiv}

\usepackage[utf8]{inputenc} 
\usepackage[T1]{fontenc}    
\usepackage{lmodern}        
\usepackage{hyperref}       
\usepackage{url}            
\usepackage{booktabs}       
\usepackage{amsfonts}       
\usepackage{nicefrac}       
\usepackage{microtype}      
\usepackage{graphicx}

\title{latrend: A Framework for Clustering Longitudinal Data}

\author{
    Niek Den Teuling \orcidlink{0000-0003-1026-5080}
   \\
    Eindhoven University of Technology \\ Philips Research \\
   \\
  \texttt{\href{mailto:niek.den.teuling@philips.com}{\nolinkurl{niek.den.teuling@philips.com}}} \\
   \And
    Steffen Pauws \orcidlink{0000-0003-2257-9239}
   \\
    Tilburg University \\ Philips Research \\
   \\
  \texttt{\href{mailto:s.c.pauws@tilburguniversity.edu}{\nolinkurl{s.c.pauws@tilburguniversity.edu}}} \\
   \And
    Edwin van den Heuvel \orcidlink{0000-0001-9157-7224}
   \\
    Eindhoven University of Technology \\
   \\
  \texttt{\href{mailto:e.r.v.d.heuvel@tue.nl}{\nolinkurl{e.r.v.d.heuvel@tue.nl}}} \\
  }

\usepackage{color}
\usepackage{fancyvrb}

\DefineVerbatimEnvironment{Highlighting}{Verbatim}{commandchars=\\\{\}}
\usepackage{framed}
\definecolor{shadecolor}{RGB}{248,248,248}
\newenvironment{Shaded}{\begin{snugshade}}{\end{snugshade}}

\newcommand{\AttributeTok}[1]{\textcolor[rgb]{0.13,0.29,0.53}{#1}}

\newcommand{\ConstantTok}[1]{\textcolor[rgb]{0.56,0.35,0.01}{#1}}
\newcommand{\ControlFlowTok}[1]{\textcolor[rgb]{0.13,0.29,0.53}{\textbf{#1}}}

\newcommand{\DecValTok}[1]{\textcolor[rgb]{0.00,0.00,0.81}{#1}}

\newcommand{\FunctionTok}[1]{\textcolor[rgb]{0.13,0.29,0.53}{\textbf{#1}}}

\newcommand{\NormalTok}[1]{#1}

\newcommand{\OtherTok}[1]{\textcolor[rgb]{0.56,0.35,0.01}{#1}}

\newcommand{\SpecialCharTok}[1]{\textcolor[rgb]{0.81,0.36,0.00}{\textbf{#1}}}

\newcommand{\StringTok}[1]{\textcolor[rgb]{0.31,0.60,0.02}{#1}}

\providecommand{\tightlist}{%
  \setlength{\itemsep}{0pt}\setlength{\parskip}{0pt}}

\usepackage{longtable,booktabs,array}
\usepackage{calc} 
\usepackage{etoolbox}
\makeatletter
\patchcmd\longtable{\par}{\if@noskipsec\mbox{}\fi\par}{}{}
\makeatother
\IfFileExists{footnotehyper.sty}{\usepackage{footnotehyper}}{\usepackage{footnote}}
\makesavenoteenv{longtable}

\usepackage{orcidlink}
\usepackage[english]{babel}
\usepackage[authoryear]{natbib}
\usepackage{amsmath}
\usepackage{array}
\usepackage{subfig}

\begin{document}
\maketitle

\begin{abstract}
Clustering of longitudinal data is used to explore common trends among
subjects over time for a numeric measurement of interest. Various
\texttt{R} packages have been introduced throughout the years for
identifying clusters of longitudinal patterns, summarizing the
variability in trajectories between subject in terms of one or more
trends. We introduce the \texttt{R} package \texttt{latrend} as a
framework for the unified application of methods for longitudinal
clustering, enabling comparisons between methods with minimal coding.
The package also serves as an interface to commonly used packages for
clustering longitudinal data, including \texttt{dtwclust},
\texttt{flexmix}, \texttt{kml}, \texttt{lcmm}, \texttt{mclust},
\texttt{mixAK}, and \texttt{mixtools}. This enables researchers to
easily compare different approaches, implementations, and method
specifications. Furthermore, researchers can build upon the standard
tools provided by the framework to quickly implement new cluster
methods, enabling rapid prototyping. We demonstrate the functionality
and application of the \texttt{latrend} package on a synthetic dataset
based on the therapy adherence patterns of patients with sleep apnea.
\end{abstract}


\hypertarget{sec:intro}{%
\section{Introduction}\label{sec:intro}}

In this work, we consider the case where subjects are repeatedly
measured on the same variable over a period of time. This type of data
is referred to as longitudinal data. No two subjects are identical, and
therefore observations made across subjects may develop differently over
time. Modeling the variability between subjects leads to an improved
understanding of the different trajectories that may occur.

Usually, longitudinal datasets are represented by a single general
trend, i.e., an average representative trajectory indicating the
expected change and variability over time. However, there may be
structural deviations from the trend caused by observed and unobserved
factors, or the distribution of random deviations is difficult to model
parametrically. In both situations, multiple common trends, i.e.,
longitudinal clusters, may provide a better representation of the data
\citep{hamaker2012researchers}.

Clustering longitudinal data is a practical approach for exploring or
representing the variability between subjects in more detail. Here, the
variability is summarized in terms of a manageable number of common
trends, which are identified in an unsupervised manner from the data
using a cluster algorithm. The approach is especially useful for
exploring datasets involving a large number of trajectories, where a
visual inspection of the trajectories would be impractical. In essence,
the data is assumed to comprise several groups, each with a different
longitudinal data generating mechanism. It differs from cross-sectional
clustering due to the need to account for the dependency between
observations within subjects, and the presence of temporal correlation
of the repeated measurements.

The exploration of subgroups in longitudinal studies is of interest in
many domains, including recidivism behavior in criminology, the
development of adolescent antisocial behavior or substance use in
psychology, and medication adherence in medicine.

Another example application is for exploring the different ways in which
patients with sleep apnea adherence to positive airway pressure (PAP)
therapy over time. Here, therapy adherence is measured in terms of the
number of hours of sleep during which the therapy is used, recorded
daily. Patients exhibit different levels of adherence to the therapy,
depending on many factors such as their sleep schedule, motivation,
self-efficacy, and the perceived importance of therapy
\citep{cayanan2019review}. Moreover, patients may exhibit a different
level of change over time, depending on their initial usage and their
ability to adjust to the therapy. To account for the many possibly
unobserved factors involved, researchers have used longitudinal
clustering to summarize the between-subject variability in terms of
longitudinal patterns of therapy adherence
\citep{babbin2015identifying, denteuling2021latent, yi2022identifying}.

A number of packages have been created in \texttt{R}
\citep{rcoreteam2021r} that can be used for clustering longitudinal
data. However, for researchers analyzing a novel case study, choosing
the best method or implementation is not straightforward due to the
inherent exploratory nature of such an analysis. Considering that each
of these packages have been created to fulfill a gap in the capabilities
of other existing implementations or approaches, there is value in
comparing the results for new case studies at hand. In any case, the
evaluation of different approaches across packages is an activity of
considerable effort, as the methods, inputs, estimation procedure, and
cluster representations differ greatly between packages.

The aim of the \texttt{latrend} package is to facilitate the exploration
of heterogeneity in longitudinal datasets through a variety of cluster
methods from various fields of research in a standardized manner. The
package provides a unifying framework, enabling users to specify,
estimate, select, compare, and evaluate any supported longitudinal
cluster method in an easy and consistent way, with minimal coding. Most
importantly, users can easily compare results between different
approaches, or run a simulation study. The \texttt{latrend} package is
available from the Comprehensive R Archive Network (CRAN) at
(\url{https://CRAN.R-project.org/package=latrend}) and on GitHub at
(\url{https://github.com/philips-software/latrend}).

A second aim of the package is extensibility so that users are able to
extend the framework with new methods or add support for another
existing method by creating a new implementation of the framework
interface. The effort in implementing new methods is considerably
reduced due to the standard longitudinal cluster functionality provided
by the framework.

Currently, a total of 18 methods for longitudinal clustering are
supported. To provide support for such a variety of approaches, the
\texttt{latrend} package interfaces with an extensive set of packages
that provide methods that are applicable for clustering longitudinal
data, including \texttt{akmedoids} \citep{Adepeju2020akmedoids},
\texttt{crimCV} \citep{Nielsen2018crimCV}, \texttt{dtwclust}
\citep{sardaespinosa2019time}, \texttt{flexmix}
\citep{gruen2008flexMix}, \texttt{funFEM} \citep{Bouveyron2015funFEM},
\texttt{kml} \citep{genolini2015kml}, \texttt{lcmm}
\citep{proustlima2017estimation}, \texttt{mclust}
\citep{Scrucca2016mclust}, \texttt{mixAK} \citep{Komarek2009New}, and
\texttt{mixtools} \citep{benaglia2009mixtools}. In this way, we build
upon the cluster packages created by the \texttt{R} community. Support
has also been added for MixTVEM; a mixture model proposed and
implemented as an \texttt{R} script by \citet{dziak2015modeling}.

To the best of our knowledge, such a comprehensive package does not yet
exist in the context of clustering longitudinal data. The
\texttt{latrend} package has similar aspirations as the \texttt{flexmix}
package \citep{gruen2008flexMix}, which also provide extensible
framework for (multilevel) clustering. However, the scope of our package
is purposefully broader, to facilitate users to apply approaches from
various fields of research. Our framework is agnostic to the
specification, estimation, and representation used by the methods.

The paper is organized as follows. A short overview of different
approaches to clustering longitudinal data is given in Section
\protect\hyperlink{sec:methods}{2}. In Section
\protect\hyperlink{sec:design}{3}, the design principles and high-level
structure of the framework are described. The usage of the package is
demonstrated in Section \protect\hyperlink{sec:demo}{4}. Section
\protect\hyperlink{sec:extension}{5} describes three ways in which users
can implement their own cluster methods. Lastly, a summary and future
steps are presented in Section \protect\hyperlink{sec:discussion}{6}.

\hypertarget{sec:methods}{%
\section{Methods}\label{sec:methods}}

We will briefly describe common general approaches to clustering
longitudinal data. Moreover, we summarize the main strengths of these
approaches. For brevity, we do not go into the specifics of any the
packages. We refer to the accompanying articles of these packages for
further details.

We begin by describing the aspects that all of the approaches have in
common. Let the repeated observations of the trajectory from subject
\(i\) be denoted by \[\mathbf{y}_i = (y_{i1}, y_{i2}, ..., y_{iJ_i}),\]
where \(y_{ij}\) is a numerical value of some variable of interest,
\(t_{ij}\) is the measurement time, and \(J_i\) is the number of
observations of trajectory \(\mathbf{y}_i\) for subject \(i\).

Regardless of the approach, any method for clustering longitudinal data
approximates the dataset heterogeneity in terms of a set of \(K\)
clusters, with each cluster representing a proportion \(\pi_k\) of the
population, with \(\pi_k > 0\) and \(\sum_{k = 1}^K \pi_k = 1\). The
clusters may be discovered by identifying groupings of similar subjects,
based on their trajectory. Typically, a cluster method is estimated for
a given number of clusters, specified by the user. By applying a cluster
method for a different number of clusters, the most appropriate number
of clusters can then be determined for the respective data.

Subjects are generally assumed to belong to a single cluster. Therefore,
many cluster methods partition the subjects into \(k\) mutually
exclusive sets \(I_1, I_2, ..., I_K\), where \(I_k\) denotes the set of
subjects to belong to cluster \(k\), with \(\bigcup_{k=1}^{K}I_{k}=I\).
Depending on the application, it may be desirable to identify a
representation for each cluster, also referred to as the cluster center,
which provides a summary of the cluster. This representation may be
obtained from the averaged representation of all the subjects assigned
to the respective cluster, by designating a representative subject, or
through the cluster representation defined by the method, if applicable.

Other cluster methods allow for overlapping clusters, commonly referred
to as soft or fuzzy clustering. Here, subjects may belong to multiple
clusters, with a certain degree or weight to which subjects belong to
each cluster. In the case of model-based clustering
\citep{mcnicholas2010model}, the clusters are represented by a mixture
of statistical models, for which cluster membership is expressed as a
probability. In applications where each subject is assumed to belong to
one cluster, subjects are typically assigned to the cluster with the
highest subject-specific posterior probability, referred to as modal
assignment.

\hypertarget{cross-sectional-clustering}{%
\subsection{Cross-sectional
clustering}\label{cross-sectional-clustering}}

In a cross-sectional cluster approach, also referred to as a
raw-data-based approach \citep{liao2005clustering}, the different
observation moments are treated as separate features for a standard
cluster algorithm, i.e., as if we are conducting a cross-sectional
cluster analysis. In standard cluster algorithms such as \(k\)-means,
the features are assumed to be independent, although this is generally
not a strict requirement. The temporal independence assumption made in
this approach yields a non-parametric representation of the
trajectories. This makes it a useful approach for an exploratory
analysis without any prior assumptions on the shape of the trajectories.
The main limitation of this approach is that observations must be
aligned between trajectories, i.e., measured at the same respective
moments in time. Consequently, missing observations should be imputed.

An example of a cross-sectional approach is longitudinal \(k\)-means
(KmL). KmL applies the \(k\)-means cluster algorithm directly to the
observations. The cluster trajectories are determined by the averaged
observations of trajectories assigned to the respective cluster. The
method is implemented in the \texttt{kml} package by
\citet{genolini2015kml}.

A model-based cross-sectional approach is seen in longitudinal latent
profile analysis (LLPA), otherwise known as longitudinal latent class
analysis \citep{muthen2004latent}. Here, latent profile analysis, more
commonly referred to as a Gaussian mixture model, is used to describe
each moment in time as a normally distributed random variable. A dataset
with trajectories each comprising \(J\) observations is thus described
by \(J\) independent normals, each modeling the response distribution at
a different moment in time. Gaussian mixture models, and thereby LLPA,
can be estimated using, for example, the \texttt{mclust} package by
\citet{Scrucca2016mclust}.

\hypertarget{distance-based-clustering}{%
\subsection{Distance-based clustering}\label{distance-based-clustering}}

Distance-based cluster algorithms operate on the pairwise distance
between trajectories. Such methods take a distance matrix as input,
where the choice of the distance metric, i.e., the dissimilarity
measure, is left to the user. Examples of cluster algorithms that use
this approach include \(k\)-medoids and agglomerative hierarchical
clustering.

Given the trajectories of subject \(a\) and \(b\), the distance metric
is denoted by \(d(\mathbf{y}_a, \mathbf{y}_b)\). As an example, the
Euclidean distance
\[d(\mathbf{y}_a, \mathbf{y}_b) = \sqrt{\sum_{j} (y_{bj} - y_{aj})^2}.\]
may be used as the distance metric. Cross-sectional clustering is a
special case of distance-based clustering where a raw-data distance
metric is used.

The approach is commonly used for time series
clustering\footnote{Clustering longitudinal data can be regarded as a special case of time series clustering where the time series have a common starting point.},
and the list of available distance metrics that have been proposed over
the past decades is extensive \citep{aghabozorgi2015time}. A distance
function can be specified to account for one or more temporal aspects of
interest, e.g., mean level, changes over time, variability,
autocorrelation, spectral components, and entropy. Many dissimilarity
metrics are implemented in the \texttt{dtwclust} package
\citep{sardaespinosa2019time}.

\hypertarget{regression-based-clustering}{%
\subsection{Regression-based
clustering}\label{regression-based-clustering}}

In regression-based clustering, the longitudinal dataset is modeled by a
regression model comprising a mixture of submodels \citep{de2008model}.
It is also referred to as latent-class trajectory modeling. This
approach comprises a versatile class of (semi-)parametric methods. Most
importantly, the shape of the trajectories can be represented using a
parametric model, requiring fewer parameters compared to a
non-parametric approach. Measurements can be taken at different times
between subjects, and covariates can be accounted for. Moreover, users
can incorporate assumptions into the modeling of the trajectories and
clusters, such as the distribution of the response variable, the
within-cluster variability, and heteroskedasticity.

A straightforward example of regression-based clustering involves
modeling the population as a mixture of cluster trajectory models. This
is referred to as group-based trajectory modeling (GBTM) or latent-class
growth analysis (LCGA). It is essentially a mixture of linear regression
models, with \begin{equation} \label{eqn:gbtm_y}
  y_{ij} = \mathbf{x}_{ij} \mathbf{\beta}_k + \varepsilon_{ijk} \quad \textrm{for } i \in I_k,
\end{equation} where \(\mathbf{x}_{ij}\) is the \(N \times B\) design
matrix of \(B\) covariates, \(\mathbf{\beta}_k\) are the \(B\)
group-specific coefficients, and \(\varepsilon_{ijk}\) is the normally
distributed residual error with zero mean and constant variance
\(\sigma_k^2\) which may be specified to differ between clusters. The
design matrix contains covariates of time, enabling the model to
describe the change in response over time. External covariates can be
included to further explain the dependent variable. The expected values
of a trajectory, assuming the trajectory belongs to cluster \(k\), is
given by \begin{equation}
  \label{eqn:gbtm}
  E(y_{ij} | C_i = k) = \mathbf{x}_{ij} \mathbf{\beta}_k.
\end{equation} GBTM is available, for example, in the packages
\texttt{lcmm} \citep{proustlima2017estimation} and \texttt{crimCV}
\citep{Nielsen2018crimCV}.

A popular form of regression-based clustering that does consider
within-cluster variability is growth mixture modeling (GMM)
\citep{muthen2004latent}, which represents a mixture of multilevel
models. Here, the within-cluster variability is modeled by allowing for
subject-specific deviations from the cluster center, e.g., a deviation
in the intercept. Using a linear mixed modeling approach, the
trajectories for cluster \(k\) are given by \begin{equation}
  y_{ij} = \mathbf{x}_{ij} \mathbf{\beta}_k + \mathbf{z}_{ij} \mathbf{u}_{ki} + \varepsilon_{ijk} \quad \textrm{for } i \in I_k.
\end{equation} Here, \(\mathbf{z}_{ij}\) is the \(N \times U\) design
matrix for the \(U\) random effects, and \(\mathbf{u}_{ki}\) are the
subject-specific random coefficients for cluster \(k\). The random
effects are assumed to be normally distributed with mean zero and
variance-covariance matrix \(\Sigma_k\). The expected values of a
trajectory, assuming the trajectory belongs to cluster \(k\), is given
by \begin{equation}
  \label{eqn:gmm}
  E(y_{ij} | C_i = k, \mathbf{u}_{i}) = \mathbf{x}_{ij} \mathbf{\beta}_k + \mathbf{z}_{ij} \mathbf{u}_{ki}.
\end{equation} GMM is available in packages such as \texttt{lcmm}
\citep{proustlima2017estimation}, \texttt{mixtools}
\citep{benaglia2009mixtools}, and \texttt{mixAK} \citep{Komarek2009New}.

A challenge with this approach is the many parameters that need to be
estimated, which typically increases linearly with the number of
clusters. The estimation may fail to converge or may yield empty
clusters\footnote{Empty clusters occur when no trajectories are most likely to belong to the respective cluste, i.e., the cluster has no trajectories under modal assignment}.
This is usually handled by repeatedly fitting the model with random
starts, or by providing better starting values for the coefficients.

\hypertarget{feature-based-clustering}{%
\subsection{Feature-based clustering}\label{feature-based-clustering}}

In a feature-based approach, each trajectory is independently
represented by a set of temporal characteristics (i.e., features,
coefficients), for example, the mean, variability, and change over time
\citep{liao2005clustering}. The trajectories are then clustered based on
the features or coefficients using a cross-sectional cluster algorithm.
This can be regarded as a special case of distance-based clustering, but
with a domain-tailored distance function. This approach has the
advantage of allowing users to easily combine arbitrary features of
interest. The approach is used, for example, by the anchored
\(k\)-medoids algorithm provided by the \texttt{akmedoids} package
\citep{Adepeju2020akmedoids}. Here, the trajectories are represented
using linear regression models, and are clustered based on the model
coefficients.

Compared to the rather time-intensive regression-based clustering
approach, the trajectory models only need to be estimated once. A
disadvantage compared to regression-based clustering is that the
reliability of the trajectory coefficients depends on the available data
per trajectory. This approach therefore generally requires a greater
number of observations per subject to yield similar results.

\hypertarget{identifying-the-number-of-clusters}{%
\subsection{Identifying the number of
clusters}\label{identifying-the-number-of-clusters}}

Due to the exploratory nature of clustering, the number of clusters is
typically not known. Moreover, most of the cluster methods require the
user to specify the number of clusters. The preferred number of clusters
for the respective method can be determined by estimating the method for
an increasing number of clusters, followed by comparing the solutions by
means of an evaluation metric. In such a comparison for a particular
method, the interpretation of the metric is consistent across the
solutions, as they all originate from the same method specification.

Many metrics are available, depending on the type of method that is
being applied. For example, in distance-based methods, the solutions are
typically evaluated in terms of the separation between clusters. Cluster
separation is measured by the distance between trajectories or cluster
trajectories, e.g., using the average Silhouette width (ASW)
\citep{rousseeuw1987silhouettes} or the Dunn index
\citep{arbelaitz2013extensive}. In contrast, a regression-based approach
typically has no notion of the distance between trajectories, but
instead measures the likelihood of the overall regression model on the
given the data, enabling the use of likelihood-based evaluation such as
the Bayesian information criterion (BIC), Akaike information criterion
(AIC), or likelihood ratio test \citep{vandernest2020overview}. Specific
to cluster regression methods where the longitudinal observations are
modeled at the subject level, assessing the solution in terms of the
residual errors of the trajectories may be of interest. Examples of such
metrics include the mean absolute error (MAE) and root mean squared
error (RMSE). For probabilistic assignments these metrics may be
weighted by the posterior probability of the trajectories, denoted by
WMAE and WRMSE, respectively.

Overall, the preferred metric depends on the type of method under
consideration and the case study domain. Users are advised to follow
recommendations from literature for the respective method. Moreover, it
is advisable to use the evaluation metric merely as guidance in
identifying the preferred solution, as a trade-off between the number of
clusters and the interpretability of the solution. Lastly, it is
worthwhile to factor in domain knowledge into the selection of cluster
solutions \citep{nagin2018group}.

\hypertarget{methods-compare}{%
\subsection{Comparing methods}\label{methods-compare}}

The approaches may yield considerably different results, arising from
fundamental differences in the temporal representation and similarity
criterion of the methods. We provide a high-level summary of strengths
and limitations of the approaches in Table \ref{tbl:compare}, which
helps to guide the user towards an initial selection of applicable
approaches relative to the case study at hand. Note that even for
methods of the same type of approach, results may differ depending on
how the trajectories are represented, trajectory similarity is measured,
or how clusters are formed. Considering that the most suitable approach
or method is typically not known in advance, it is advisable to evaluate
and compare the solutions between methods to identify the most suitable
method for the respective case study. The resulting solutions can then
be compared using an external evaluation metric.

A useful starting point in comparing the preferred solutions between
methods is to evaluate the similarity between the cluster partitions.
After all, if both candidate methods find a similar cluster partition,
this would indicate that both methods find the same grouping despite
representational differences. In contrast, if the cluster partitions are
dissimilar, it may suggest that either a hybrid approach could be of
interest, or that one method is preferred over the other.

The similarity between cluster partitions of two methods can be assessed
using partition similarity metrics such as the adjusted Rand index (ARI)
\citep{hubert1985comparing}, variance of information, or the split-join
index. These metrics are applicable to any method and are even
applicable when the solutions have a mismatching number of clusters. In
some case studies, a ground truth may be available in the form of a
reference cluster partition. Partition similarity metrics such as the
ARI may then be used to identify the solution that most closely
resembles the ground truth. Alternatively, users may obtain a partial
ground truth by manually annotating a subset of the trajectories based
on domain knowledge.

For methods that have a longitudinal representation of the clusters, it
can be insightful to assess the similarity between the cluster
trajectories of one solution to another (reference) solution. A possible
metric for this is the weighted minimum mean absolute error (WMMAE)
\citep{denteuling2021comparison}, which evaluates the WMAE for each
cluster trajectory \(\hat{\mathbf{y}}_{k}\) with its nearest reference
cluster \(\hat{y}_{kj}\). It is defined as
\begin{equation}\label{eqn:wmmae}
\textrm{WMMAE} = \frac{1}{J} \sum_{k=1}^K \left[ \pi_k \min_{k^\prime \in \{1,\ldots,K\}} \sum_{j=1}^J \mathopen| y_{k j} - \hat{y}_{k^ \prime j} \mathclose| \right],
\end{equation} where \(J\) is the number of moments in time on which the
cluster trajectories are compared to the reference cluster trajectories.
Here, \(y_{kj}\) and \(\hat{y}_{kj}\) denote the expected value of the
cluster trajectory at time \(t_j\) for the solution and the reference
solution, respectively. The interpretation of the value of the WMMAE is
relative to the scale of the response variable.

Solutions may be compared further by assessing the compactness of the
clusters or the separation between clusters on a common distance metric,
for example using the average Silhouette width or the Dunn index. This
is useful to identify the method that is best at identifying distinct
subgroups.

\begin{longtable}[]{@{}
  >{\raggedright\arraybackslash}p{(\columnwidth - 4\tabcolsep) * \real{0.0936}}
  >{\raggedright\arraybackslash}p{(\columnwidth - 4\tabcolsep) * \real{0.4483}}
  >{\raggedright\arraybackslash}p{(\columnwidth - 4\tabcolsep) * \real{0.4532}}@{}}
\caption{Summary of the general strengths of limitations of the
different approaches to longitudinal clustering.
\label{tbl:compare}}\tabularnewline
\toprule\noalign{}
\begin{minipage}[b]{\linewidth}\raggedright
Approach
\end{minipage} & \begin{minipage}[b]{\linewidth}\raggedright
Strengths
\end{minipage} & \begin{minipage}[b]{\linewidth}\raggedright
Limitations
\end{minipage} \\
\midrule\noalign{}
\endfirsthead
\toprule\noalign{}
\begin{minipage}[b]{\linewidth}\raggedright
Approach
\end{minipage} & \begin{minipage}[b]{\linewidth}\raggedright
Strengths
\end{minipage} & \begin{minipage}[b]{\linewidth}\raggedright
Limitations
\end{minipage} \\
\midrule\noalign{}
\endhead
\bottomrule\noalign{}
\endlastfoot
Cross-sectional & \begin{minipage}[t]{\linewidth}\raggedright
\begin{itemize}
\tightlist
\item
  No assumptions on the shape of the cluster trajectories
\item
  Low sample size requirement
\item
  Very fast to estimate
\item
  Suitable for initial exploration
\end{itemize}
\end{minipage} & \begin{minipage}[t]{\linewidth}\raggedright
\begin{itemize}
\tightlist
\item
  Requires time-aligned trajectories of equal length
\item
  Requires complete data
\item
  Does not account for the temporal relation of observations
\end{itemize}
\end{minipage} \\
Distance-based & \begin{minipage}[t]{\linewidth}\raggedright
\begin{itemize}
\tightlist
\item
  Flexible in the choice of distance metric(s)
\item
  Trajectory distance matrix only needs to be computed once
\item
  Fast to estimate
\end{itemize}
\end{minipage} & \begin{minipage}[t]{\linewidth}\raggedright
\begin{itemize}
\tightlist
\item
  Distance matrix computation is not practical for a large number of
  trajectories
\item
  Pairwise comparison of trajectories is more sensitive to noise
\item
  Many distance metrics require time-aligned trajectories
\end{itemize}
\end{minipage} \\
Regression-based & \begin{minipage}[t]{\linewidth}\raggedright
\begin{itemize}
\tightlist
\item
  Low sample size requirements due to inclusion of parametric
  assumptions
\item
  Can handle missing data
\item
  Can handle trajectories of unequal length and variable time
\item
  Can account for covariates
\item
  Relatively robust to trajectories that do not fit the representation
\end{itemize}
\end{minipage} & \begin{minipage}[t]{\linewidth}\raggedright
\begin{itemize}
\tightlist
\item
  May be challenging to estimate (convergence problems)
\item
  Computationally intensive to estimate
\end{itemize}
\end{minipage} \\
Feature-based & \begin{minipage}[t]{\linewidth}\raggedright
\begin{itemize}
\tightlist
\item
  Temporal features only needs to be computed once
\item
  Very fast to estimate
\item
  Fast alternative to regression-based approach given a sufficiently
  large sample size
\end{itemize}
\end{minipage} & \begin{minipage}[t]{\linewidth}\raggedright
\begin{itemize}
\tightlist
\item
  Sensitive to trajectories that do not fit the representation
\item
  Trajectory-independent feature estimation is more sensitive to
  observational outliers
\end{itemize}
\end{minipage} \\
\end{longtable}

\hypertarget{sec:design}{%
\section{Software design}\label{sec:design}}

We begin by providing a high-level description of the framework,
outlining the main functionality of the classes. A step-by-step
demonstration of the framework is given in the next section. The
software is built on an object-oriented paradigm using the S4 system,
available in the \texttt{methods} package \citep{rcoreteam2021r}. We
have chosen to use the S4 paradigm over S3 due to the support for class
inheritance, object validation, and method signatures.

The framework is designed to provide a standardized way of specifying,
estimating, and evaluating different longitudinal cluster methods. This
is achieved by defining two interfaces: the \texttt{lcMethod} interface
is used for defining the specification and estimation logic of a method.
The \texttt{lcModel} interface represents the result of an estimated
method. Using these two interfaces, we can then define method-agnostic
estimation procedures for applying a specified method to a given
dataset, yielding a method result. This estimation procedure is
implemented by the \texttt{latrend()} function. For example, users can
specify a growth mixture model (GMM) through a \texttt{lcMethodLcmmGMM}
object, specifying the GMM and the estimation settings. The resulting
estimated GMM is represented by a \texttt{lcModelLcmmGMM} object.

A key advantage of having stand-alone estimation procedures is that it
enables standard validation of the inputs and outputs, which otherwise
would need to be implemented for each method. Moreover, it ensures all
methods take the same data format as input, and allows for more
procedures to be implemented which automatically support all implemented
methods. There are additional procedures implemented in the package,
including repeated estimation via \texttt{latrendRep()}, batch
estimation via \texttt{latrendBatch()}, and bootstrap sampling via
\texttt{latrendBoot()}.

\hypertarget{dataset-input}{%
\subsection{Dataset input}\label{dataset-input}}

We have selected the \texttt{data.frame} in long format as the preferred
representation for longitudinal datasets. Here, each row represents an
observation for a trajectory at a given time, possibly for multiple
covariates. The trajectory and time of an observation are indicated in
separate columns. This format can represent irregularly timed
measurements, a variable number of observations per trajectory, and an
arbitrary number of covariates of different types. Since not all
datasets are readily available in this format, the \texttt{latrend()}
estimation procedures handle data input by calling the generic
\texttt{transformLatrendData()} function. Currently, this transformation
is only defined for \texttt{matrix} input. Users can implement the
method to add support for other longitudinal data types.

\hypertarget{subsec:lcmethod}{%
\subsection{\texorpdfstring{The \texttt{lcMethod}
class}{The lcMethod class}}\label{subsec:lcmethod}}

The \texttt{lcMethod} class has two purposes. The first purpose is to
record the method specification, defined by the method parameters and
other settings, referred to as the method arguments. The second purpose
is to provide the logic for estimating the method for the specified
arguments and given data. \texttt{lcMethod} objects are immutable. Users
only interact with a \texttt{lcMethod} object for retrieving method
arguments, or for creating a new specification with modified arguments.
This functionality is provided by the base \texttt{lcMethod} class.

The base class also stores the method arguments in a list, inside the
\texttt{arguments} slot. The method arguments can be of any type. The
names of subclasses are prefixed by ``\emph{lcMethod}''. Subclasses can
validate the model arguments against the data by overriding the
\texttt{validate()} function. Due to the specific internal structure of
a \texttt{lcMethod} object, constructors are defined for creating
\texttt{lcMethod} objects of a specific class for a given set of
arguments. In \texttt{lcMethod} implementations that are a wrapper
around an existing cluster package function, the method arguments are
simply passed to the package function. The required arguments and their
default values are obtained from the formal function arguments of the
package function at runtime.

The evaluation of the method arguments is delayed until the method
estimation process. This enables a \texttt{lcMethod} object to be
printed in an easily readable way, where the original argument
expressions or calls are shown, instead of the evaluation result. This
is useful when an argument takes on a function or complex data
structure, and it reduces the memory footprint when a large set of
method permutations is generated and serialized, such as in a simulation
study.

The method estimation process is implemented through six generic
functions: \texttt{prepareData()}, \texttt{compose()},
\texttt{validate()}, \texttt{preFit()}, \texttt{fit()}, and
\texttt{postFit()}. The purpose of each step is explained in Section
\protect\hyperlink{sec:extension}{5}. There are several advantages to
this design. Firstly, the structure enables the method estimation
process to be checked at each step. Secondly, splitting the estimation
logic into processing steps encourages shorter functions with clearer
functionality, resulting in more readable code. Thirdly, the steps
enable optimizations in the case of repeated method estimation, for
which the \texttt{prepareData()} function only needs to be called once.
Lastly, in case of an update to the \texttt{lcModel} post-processing
step, the \texttt{postFit()} function can be applied to previously
obtained \texttt{lcModel} objects.

\hypertarget{supported-methods}{%
\subsubsection{Supported methods}\label{supported-methods}}

An overview of the currently available methods that can be specified is
given in Table \ref{tbl:methods}. The \texttt{lcMethodGCKM} class
implements a feature-based approach, based on representing the
trajectories through a linear mixed model specified in the \texttt{lme4}
package \citep{Bates2015Fitting}.

\begin{longtable}[]{@{}
  >{\raggedright\arraybackslash}p{(\columnwidth - 4\tabcolsep) * \real{0.1880}}
  >{\raggedright\arraybackslash}p{(\columnwidth - 4\tabcolsep) * \real{0.5188}}
  >{\raggedright\arraybackslash}p{(\columnwidth - 4\tabcolsep) * \real{0.2857}}@{}}
\caption{The list of currently supported methods for clustering
longitudinal data, in alphabetical order. The methods in the bottom row
represent generic approaches which can be adapted.
\label{tbl:methods}}\tabularnewline
\toprule\noalign{}
\begin{minipage}[b]{\linewidth}\raggedright
Class
\end{minipage} & \begin{minipage}[b]{\linewidth}\raggedright
Method
\end{minipage} & \begin{minipage}[b]{\linewidth}\raggedright
Package
\end{minipage} \\
\midrule\noalign{}
\endfirsthead
\toprule\noalign{}
\begin{minipage}[b]{\linewidth}\raggedright
Class
\end{minipage} & \begin{minipage}[b]{\linewidth}\raggedright
Method
\end{minipage} & \begin{minipage}[b]{\linewidth}\raggedright
Package
\end{minipage} \\
\midrule\noalign{}
\endhead
\bottomrule\noalign{}
\endlastfoot
\texttt{lcMethodAkmedoids} & Anchored \emph{k}-medoids &
\texttt{akmedoids} \citep{Adepeju2020akmedoids} \\
\texttt{lcMethodCrimCV} & Group-based trajectory modeling of count data
& \texttt{crimCV} \citep{Nielsen2018crimCV} \\
\texttt{lcMethodDtwclust} & Dynamic time warping & \texttt{dtwclust}
\citep{sardaespinosa2019time} \\
\texttt{lcMethodFlexmix} & Interface to FlexMix framework &
\texttt{flexmix} \citep{gruen2008flexMix} \\
\texttt{lcMethodFlexmixGBTM} & Group-based trajectory modeling &
\texttt{flexmix} \citep{gruen2008flexMix} \\
\texttt{lcMethodFunFEM} & funFEM & \texttt{funFEM}
\citep{Bouveyron2015funFEM} \\
\texttt{lcMethodGCKM} & Feature-based clustering using growth curve
modeling and \emph{k}-means & \texttt{lme4} \citep{Bates2015Fitting} \\
\texttt{lcMethodKML} & longitudinal \emph{k}-means & \texttt{kml}
\citep{genolini2015kml} \\
\texttt{lcMethodLcmmGBTM} & Group-based trajectory modeling &
\texttt{lcmm} \citep{proustlima2017estimation} \\
\texttt{lcMethodLcmmGMM} & Growth mixture modeling & \texttt{lcmm}
\citep{proustlima2017estimation} \\
\texttt{lcMethodLMKM} & Feature-based clustering using linear regression
and \emph{k}-means & \\
\texttt{lcMethodMclustLLPA} & Longitudinal latent profile analysis &
\texttt{mclust} \citep{Scrucca2016mclust} \\
\texttt{lcMethodMixAK\_GLMM} & Mixture of generalized linear mixed
models & \texttt{mixAK} \citep{Komarek2009New} \\
\texttt{lcMethodMixtoolsGMM} & Growth mixture modeling &
\texttt{mixtools} \citep{benaglia2009mixtools} \\
\texttt{lcMethodMixtoolsNPRM} & Non-parametric repeated measures
clustering & \texttt{mixtools} \citep{benaglia2009mixtools} \\
\texttt{lcMethodMixTVEM} & Mixture of time-varying effects models & \\
\texttt{lcMethodRandom} & Random partitioning & \\
\texttt{lcMethodStratify} & Stratification rule & \\
\texttt{lcMethodFeature} & Feature-based clustering & \\
\end{longtable}

Additionally, a partitioning of trajectories can be specified without an
estimation step through the \texttt{lcModelPartition} and
\texttt{lcModelWeightedPartition} classes, providing trajectories with a
cluster membership or membership weight, respectively.

\hypertarget{subsec:lcmodel}{%
\subsection{\texorpdfstring{The \texttt{lcModel}
class}{The lcModel class}}\label{subsec:lcmodel}}

The \texttt{lcModel} class represents the estimated cluster solution. It
is designed to function as any other model fitted in \texttt{R}. Here,
the word ``model'' should be taken in the broadest sense of the word,
where any resulting cluster partitioning represents the data, and
thereby is regarded as a model of said data. Users can apply the
familiar functions from the \texttt{stats} package
\citep{rcoreteam2021r} where applicable, including the
\texttt{predict()}, \texttt{plot()}, \texttt{summary()},
\texttt{fitted()}, and \texttt{residuals()} functions. Furthermore,
\texttt{lcModel} objects support functions for obtaining the cluster
representation, such as the cluster proportions, sizes, names, and
trajectories.

The base \texttt{lcModel} class facilitates basic functionality such as
providing a solution summary and providing functionality for computing
predictions or fitted values. The two most important functions that
characterize the class are the \texttt{predict()} and
\texttt{postprob()} functions. These functions are used to derive the
cluster trajectories, the posterior probabilities of the trajectories,
and cluster proportions.

The base class stores information regarding the model, including the
estimated \texttt{lcMethod} object, the \texttt{call} that was used to
estimate the method, the date and time when the method was estimated,
the total estimation time, and a text label for differentiating
solutions. Users should not update the slots of the base class directly,
except for the \texttt{tag} slot, which is intended as a convenient way
of assigning custom meta data to the \texttt{lcModel}.

The names of subclasses are prefixed by ``\emph{lcModel}''. Subclasses
generally have little need for adding new slots, as most of the
functionality resides inside the class functions, such that results and
statistics are computed dynamically. This enables fitted
\texttt{lcModel} objects to be modified retroactively, e.g., for
correcting implementation errors that are discovered at a later stage.

In the \texttt{lcModel} subclass implementations that are based on an
underlying \texttt{R} package, the subclass serves as a wrapper around
the underlying package model. The underlying model is exposed via the
\texttt{getModel()} function so that users can still benefit from the
specialized functionality provided by the underlying package.

\hypertarget{subsec:metric}{%
\subsection{The metric interfaces}\label{subsec:metric}}

There is a vast number of metrics available in literature. To provide
access to as many metrics as possible, and to enable users to add
missing metrics as needed, we define an interface for the computation of
metrics. Users can replace or extend the metrics with custom
implementations. To ensure a consistent output across all metrics, the
output of metric functions must be scalar. Currently, the framework
supports any of the applicable metrics from the packages
\texttt{clusterCrit} \citep{Desgraupes2018clusterCrit} and
\texttt{mclustcomp} \citep{You2018mclustcomp}. The list of supported
internal and external metrics is obtained via the
\texttt{getInternalMetricNames()} and \texttt{getExternalMetricNames()}
functions, respectively. Metrics can be added or updated via the
\texttt{defineInternalMetric()} and \texttt{defineExternalMetric()}
functions.

\hypertarget{sec:demo}{%
\section{Using the package}\label{sec:demo}}

We illustrate the main capabilities of the package through a
step-by-step exploratory cluster analysis on the longitudinal dataset
named \texttt{PAP.adh} which is included with the package. This
synthetic dataset was simulated based on the real-world study reported
by \citet{yi2022identifying}, who investigated the longitudinal CPAP
therapy usage patterns of patients with obstructive sleep apnea since
the start of their treatment. They identified three distinct patterns of
therapy adherence: patients who were adherent to the therapy and stable
in their usage (``Adherent''), patients who were consistently
non-adherent (``Non-adherent''), and patients who improved their usage
over time (``Improvers''). We used the growth mixture model fit reported
by the authors to simulate new patients, yielding the \texttt{PAP.adh}
dataset.

The goal of the analysis is to identify the common patterns of adherence
and to establish the most suitable method for the data out of those
considered. For brevity, the description of the package function
arguments used in the demonstration below is limited to the main
arguments. We refer users to the package documentation to learn more
about other optional arguments.

The \texttt{PAP.adh} dataset comprises records of the weekly average
hours of therapy usage of 301 patients in their first 13 weeks of
therapy. Therapy usage ranges between 0 and 9.5 hours, with a mean of
4.5 hours. The \texttt{PAP.adh} dataset is represented by a
\texttt{data.frame} in long format, with each row representing the
observation of a patient at a specific week (1 to 13).

\begin{Shaded}
\begin{Highlighting}[]
\FunctionTok{library}\NormalTok{(}\StringTok{"latrend"}\NormalTok{)}
\FunctionTok{data}\NormalTok{(}\StringTok{"PAP.adh"}\NormalTok{)}
\FunctionTok{head}\NormalTok{(PAP.adh)}
\end{Highlighting}
\end{Shaded}

\begin{verbatim}
##   Patient Week UsageHours    Group
## 1       1    1   6.298703 Adherers
## 2       1    2   5.916080 Adherers
## 3       1    3   5.022241 Adherers
## 4       1    4   5.788624 Adherers
## 5       1    5   4.758154 Adherers
## 6       1    6   4.222821 Adherers
\end{verbatim}

The \texttt{Patient} column indicates the trajectory to which the
observation belongs. The \texttt{UsageHours} column represents the
averaged hours of usage in the respective therapy week, denoted by the
\texttt{Week} column. The true cluster membership per trajectory is
indicated by the \texttt{Group} column.

Throughout the analysis, there are several occasions during which the
trajectory identifier and time columns would need to be specified.
Instead of passing the column names to each function, we can set the
default index columns using the \texttt{options} mechanism. Keep in mind
that this is only recommended during interactive use.

\begin{Shaded}
\begin{Highlighting}[]
\FunctionTok{options}\NormalTok{(}\AttributeTok{latrend.id =} \StringTok{"Patient"}\NormalTok{, }\AttributeTok{latrend.time =} \StringTok{"Week"}\NormalTok{)}
\end{Highlighting}
\end{Shaded}

\begin{figure}

{\centering \includegraphics{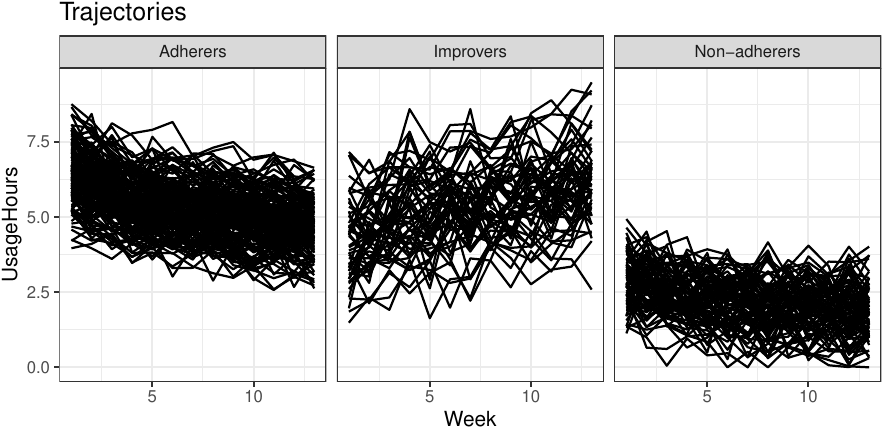} 

}

\caption{The trajectories from the `PAP.adh` dataset, by reference group.}\label{fig:data}
\end{figure}

We can visualize the patient trajectories using the
\texttt{plotTrajectories()} function, shown in Figure \ref{fig:data}. As
the ground truth is known in our synthetic example, we specified the
cluster membership of the trajectories via the \texttt{cluster}
argument, resulting in a stratified visualization.

\begin{Shaded}
\begin{Highlighting}[]
\FunctionTok{plotTrajectories}\NormalTok{(PAP.adh, }\AttributeTok{response =} \StringTok{"UsageHours"}\NormalTok{, }\AttributeTok{cluster =} \StringTok{"Group"}\NormalTok{)}
\end{Highlighting}
\end{Shaded}

\hypertarget{specifying-methods}{%
\subsection{Specifying methods}\label{specifying-methods}}

We first specify the methods to be evaluated. The first method of
interest in this case study is KmL, selected for its flexibility in
identifying patterns of any shape. The KmL method is available in the
framework through the \texttt{lcMethodKML} class, which serves as a
wrapper around the \texttt{kml()} function of the \texttt{kml} package
\citep{genolini2015kml}. The KmL method is specified through the
\texttt{lcMethodKML()} constructor function.

\begin{Shaded}
\begin{Highlighting}[]
\NormalTok{kmlMethod }\OtherTok{\textless{}{-}} \FunctionTok{lcMethodKML}\NormalTok{(}\AttributeTok{response =} \StringTok{"UsageHours"}\NormalTok{, }\AttributeTok{nClusters =} \DecValTok{2}\NormalTok{)}
\NormalTok{kmlMethod}
\end{Highlighting}
\end{Shaded}

\begin{verbatim}
## lcMethodKML specifying "longitudinal k-means (KML)"
##  time:           getOption("latrend.time")
##  id:             getOption("latrend.id")
##  nClusters:      2
##  nbRedrawing:    20
##  maxIt:          200
##  imputationMethod:"copyMean"
##  distanceName:   "euclidean"
##  power:          2
##  distance:       function() {}
##  centerMethod:   meanNA
##  startingCond:   "nearlyAll"
##  nbCriterion:    1000
##  scale:          TRUE
##  response:       "UsageHours"
\end{verbatim}

Note that any unspecified arguments have been set to the default values
defined by the \texttt{kml} package. The method arguments can be
accessed using the \texttt{\$} or \texttt{{[}{[}} operator. Requested
arguments are evaluated unless disabled by the argument
\texttt{eval\ =\ FALSE}. As can be seen in the method output below, the
time index column is obtained from the \texttt{options} mechanism by
default.

\begin{Shaded}
\begin{Highlighting}[]
\NormalTok{kmlMethod}\SpecialCharTok{$}\NormalTok{time}
\end{Highlighting}
\end{Shaded}

\begin{verbatim}
## [1] "Week"
\end{verbatim}

\begin{Shaded}
\begin{Highlighting}[]
\NormalTok{kmlMethod[[}\StringTok{"time"}\NormalTok{, eval }\OtherTok{=} \ConstantTok{FALSE}\NormalTok{]]}
\end{Highlighting}
\end{Shaded}

\begin{verbatim}
## getOption("latrend.time")
\end{verbatim}

Next, we specify the other methods of interest. We use a variety of
approaches that are applicable to this type of data. We evaluate a
feature-based approach based on LMKM as implemented in
\texttt{lcMethodLMKM}, a distance-based dynamic time warping approach
via \texttt{lcMethodDtwclust} based on the \texttt{dtwclust} package,
and the regression-based approaches via the \texttt{lcMethodLcmmGBTM}
and \texttt{lcMethodLcmmGMM} methods based on the \texttt{lcmm} package
\citep{proustlima2017estimation}. We specify the distance-based approach
using dynamic time warping. For LMKM, GBTM and GMM, we model the
trajectories using an intercept and
slope\footnote{For methods supporting `formula` input, the response variable is automatically determined from the response of the formula.}.
Moreover, GBTM and GMM are specified to use a shared diagonal
variance-covariance matrix. The GMM defines a random patient intercept.

\begin{Shaded}
\begin{Highlighting}[]
\NormalTok{dtwMethod }\OtherTok{\textless{}{-}} \FunctionTok{lcMethodDtwclust}\NormalTok{(}\AttributeTok{response =} \StringTok{"UsageHours"}\NormalTok{, }\AttributeTok{distance =} \StringTok{"dtw\_basic"}\NormalTok{)}
\NormalTok{lmkmMethod }\OtherTok{\textless{}{-}} \FunctionTok{lcMethodLMKM}\NormalTok{(}\AttributeTok{formula =}\NormalTok{ UsageHours }\SpecialCharTok{\textasciitilde{}}\NormalTok{ Week)}
\NormalTok{gbtmMethod }\OtherTok{\textless{}{-}} \FunctionTok{lcMethodLcmmGBTM}\NormalTok{(}\AttributeTok{fixed =}\NormalTok{ UsageHours }\SpecialCharTok{\textasciitilde{}}\NormalTok{ Week, }
  \AttributeTok{mixture =} \SpecialCharTok{\textasciitilde{}}\NormalTok{ Week, }\AttributeTok{idiag =} \ConstantTok{TRUE}\NormalTok{)}
\NormalTok{gmmMethod }\OtherTok{\textless{}{-}} \FunctionTok{lcMethodLcmmGMM}\NormalTok{(}\AttributeTok{fixed =}\NormalTok{ UsageHours }\SpecialCharTok{\textasciitilde{}}\NormalTok{ Week, }
  \AttributeTok{mixture =} \SpecialCharTok{\textasciitilde{}}\NormalTok{ Week, }\AttributeTok{random =} \SpecialCharTok{\textasciitilde{}} \DecValTok{1}\NormalTok{, }\AttributeTok{idiag =} \ConstantTok{TRUE}\NormalTok{)}
\end{Highlighting}
\end{Shaded}

The method arguments of a \texttt{lcMethod} object cannot be modified.
Instead, a new specification is created from the existing one with the
updated method arguments. Any \texttt{lcMethod} object can be used as a
prototype for creating a new specification with new, modified, or
removed arguments using the \texttt{update()} function. As an example,
if we would like to respecify KmL to identify three clusters, this can
be done by updating the existing specification as follows:

\begin{Shaded}
\begin{Highlighting}[]
\NormalTok{kml3Method }\OtherTok{\textless{}{-}} \FunctionTok{update}\NormalTok{(kmlMethod, }\AttributeTok{nClusters =} \DecValTok{3}\NormalTok{)}
\end{Highlighting}
\end{Shaded}

As the number of clusters is generally not known in advance, we need to
fit the methods for a range of number of clusters. Generating
specifications for a series of argument values can be done via the
\texttt{lcMethods()} function, which outputs a \texttt{list} of updated
\texttt{lcMethod} objects from a given prototype. We specify each method
for up to six
clusters\footnote{Only one to four clusters were estimated for GBTM and GMM due to the relatively excessive computation time}
using:

\begin{Shaded}
\begin{Highlighting}[]
\NormalTok{kmlMethods  }\OtherTok{\textless{}{-}} \FunctionTok{lcMethods}\NormalTok{(kmlMethod,  }\AttributeTok{nClusters =} \DecValTok{1}\SpecialCharTok{:}\DecValTok{6}\NormalTok{)}
\NormalTok{lmkmMethods }\OtherTok{\textless{}{-}} \FunctionTok{lcMethods}\NormalTok{(lmkmMethod, }\AttributeTok{nClusters =} \DecValTok{1}\SpecialCharTok{:}\DecValTok{6}\NormalTok{)}
\NormalTok{dtwMethods  }\OtherTok{\textless{}{-}} \FunctionTok{lcMethods}\NormalTok{(dtwMethod,  }\AttributeTok{nClusters =} \DecValTok{2}\SpecialCharTok{:}\DecValTok{6}\NormalTok{)}
\NormalTok{gbtmMethods }\OtherTok{\textless{}{-}} \FunctionTok{lcMethods}\NormalTok{(gbtmMethod, }\AttributeTok{nClusters =} \DecValTok{1}\SpecialCharTok{:}\DecValTok{4}\NormalTok{)}
\NormalTok{gmmMethods  }\OtherTok{\textless{}{-}} \FunctionTok{lcMethods}\NormalTok{(gmmMethod,  }\AttributeTok{nClusters =} \DecValTok{1}\SpecialCharTok{:}\DecValTok{4}\NormalTok{)}
\FunctionTok{length}\NormalTok{(gmmMethods)}
\end{Highlighting}
\end{Shaded}

\begin{verbatim}
## [1] 4
\end{verbatim}

\hypertarget{fitting-methods}{%
\subsection{Fitting methods}\label{fitting-methods}}

Using the previously created method specifications, we can estimate the
methods for the \texttt{PAP.adh} data. For estimating a single method,
we can use the \texttt{latrend()} function. The function optionally
accepts an \texttt{environment} through the \texttt{envir} argument for
evaluating the method arguments within a specific environment. The
output of the function is the fitted \texttt{lcModel} object.

\begin{Shaded}
\begin{Highlighting}[]
\NormalTok{lmkm2 }\OtherTok{\textless{}{-}} \FunctionTok{latrend}\NormalTok{(lmkmMethod, }\AttributeTok{data =}\NormalTok{ PAP.adh)}
\FunctionTok{summary}\NormalTok{(lmkm2)}
\end{Highlighting}
\end{Shaded}

\begin{verbatim}
## Longitudinal cluster model using lmkm
## lcMethodLMKM specifying "lm-kmeans"
##  time:           "Week"
##  id:             "Patient"
##  nClusters:      2
##  center:         `meanNA`
##  standardize:    `scale`
##  method:         "qr"
##  model:          TRUE
##  y:              FALSE
##  qr:             TRUE
##  singular.ok:    TRUE
##  contrasts:      NULL
##  iter.max:       10
##  nstart:         1
##  algorithm:      `c("Hartigan-Wong", "Lloyd", "Forgy", "M
##  formula:        UsageHours ~ Week
## 
## Cluster sizes (K=2):
##           A           B 
## 135 (44.9%) 166 (55.1%) 
## 
## Number of obs: 3913, strata (Patient): 301
## 
## Scaled residuals:
##     Min.  1st Qu.   Median     Mean  3rd Qu.     Max. 
## -2.52815 -0.67127 -0.06772  0.00000  0.54587  4.04438
\end{verbatim}

Instead of needing to update a method prior to calling
\texttt{latrend()}, the arguments to be updated can also be passed
directly to \texttt{latrend()}. Here, we estimate the LMKM method for
three clusters.

\begin{Shaded}
\begin{Highlighting}[]
\NormalTok{lmkm3 }\OtherTok{\textless{}{-}} \FunctionTok{latrend}\NormalTok{(lmkmMethod, }\AttributeTok{nClusters =} \DecValTok{3}\NormalTok{, }\AttributeTok{data =}\NormalTok{ PAP.adh)}
\end{Highlighting}
\end{Shaded}

Alternatively, we can achieve the same result by updating the previously
estimated two-cluster solution.

\begin{Shaded}
\begin{Highlighting}[]
\NormalTok{lmkm3 }\OtherTok{\textless{}{-}} \FunctionTok{update}\NormalTok{(lmkm2, }\AttributeTok{nClusters =} \DecValTok{3}\NormalTok{)}
\end{Highlighting}
\end{Shaded}

\hypertarget{batch-estimation}{%
\subsubsection{Batch estimation}\label{batch-estimation}}

The \texttt{latrendBatch()} function estimates a list of method
specifications. This is useful for evaluating a method for a range of
number of clusters, as we have defined above using the
\texttt{lcMethods()} function. Another use case is the improvement of
model convergence and the estimation time by tuning the control
parameters. Optimizing such parameters may yield considerably improved
convergence or considerably reduced estimation time on larger datasets.
Many of the methods have settings for the number of random starts,
maximum number of iterations, and convergence criteria. However, because
such control settings are specific to each method, we will not cover
this.

The inputs to the \texttt{latrendBatch()} function are a list of
\texttt{lcMethod} objects, and a list of datasets. The output is an
\texttt{lcModels} object, representing a list of the fitted
\texttt{lcModel} objects for each dataset. A seed is specified to ensure
reproducibility of the examples.

\begin{Shaded}
\begin{Highlighting}[]
\NormalTok{lmkmList }\OtherTok{\textless{}{-}} \FunctionTok{latrendBatch}\NormalTok{(lmkmMethods, }\AttributeTok{data =}\NormalTok{ PAP.adh, }\AttributeTok{seed =} \DecValTok{1}\NormalTok{)}
\NormalTok{lmkmList}
\end{Highlighting}
\end{Shaded}

\begin{verbatim}
## List of 6 lcModels with
##   .name .method       seed nClusters
## 1     1    lmkm  762473831         1
## 2     2    lmkm 1762587819         2
## 3     3    lmkm 1463113723         3
## 4     4    lmkm 1531473323         4
## 5     5    lmkm 1922000657         5
## 6     6    lmkm 1985277999         6
\end{verbatim}

When printing a \texttt{lcModels} object, the content is shown as a
table of method specifications. By default, only arguments which differ
between the models are shown. The table can also be obtained as a
\texttt{data.frame} by calling \texttt{as.data.frame()}. We now fit the
other methods in the same manner.

\begin{Shaded}
\begin{Highlighting}[]
\NormalTok{dtwList }\OtherTok{\textless{}{-}} \FunctionTok{latrendBatch}\NormalTok{(dtwMethods, }\AttributeTok{data =}\NormalTok{ PAP.adh, }\AttributeTok{seed =} \DecValTok{1}\NormalTok{)}
\end{Highlighting}
\end{Shaded}

For the repeated estimation of more computationally intensive methods,
we can speed up the process by using parallel computation. By setting
\texttt{parallel\ =\ TRUE}, the \texttt{latrendBatch()} function will
use the parallel back-end of the \texttt{foreach} package
\citep{weston2022foreach}. To make use of this functionality, we first
need to configure the parallel back-end:

\begin{Shaded}
\begin{Highlighting}[]
\NormalTok{nCores }\OtherTok{\textless{}{-}}\NormalTok{ parallel}\SpecialCharTok{::}\FunctionTok{detectCores}\NormalTok{(}\AttributeTok{logical =} \ConstantTok{FALSE}\NormalTok{)}
\ControlFlowTok{if}\NormalTok{ (.Platform}\SpecialCharTok{$}\NormalTok{OS.type }\SpecialCharTok{==} \StringTok{"windows"}\NormalTok{) \{}
\NormalTok{  doParallel}\SpecialCharTok{::}\FunctionTok{registerDoParallel}\NormalTok{(parallel}\SpecialCharTok{::}\FunctionTok{makeCluster}\NormalTok{(nCores))   }
\NormalTok{\} }\ControlFlowTok{else}\NormalTok{ \{}
\NormalTok{  doMC}\SpecialCharTok{::}\FunctionTok{registerDoMC}\NormalTok{(nCores)}
\NormalTok{\}}
\end{Highlighting}
\end{Shaded}

The methods can then be estimated in parallel using:

\begin{Shaded}
\begin{Highlighting}[]
\NormalTok{kmlList  }\OtherTok{\textless{}{-}} \FunctionTok{latrendBatch}\NormalTok{(kmlMethods,  }
  \AttributeTok{data =}\NormalTok{ PAP.adh, }\AttributeTok{parallel =} \ConstantTok{TRUE}\NormalTok{, }\AttributeTok{seed =} \DecValTok{1}\NormalTok{)}
\NormalTok{gbtmList }\OtherTok{\textless{}{-}} \FunctionTok{latrendBatch}\NormalTok{(gbtmMethods, }
  \AttributeTok{data =}\NormalTok{ PAP.adh, }\AttributeTok{parallel =} \ConstantTok{TRUE}\NormalTok{, }\AttributeTok{seed =} \DecValTok{1}\NormalTok{)}
\NormalTok{gmmList  }\OtherTok{\textless{}{-}} \FunctionTok{latrendBatch}\NormalTok{(gmmMethods,  }
  \AttributeTok{data =}\NormalTok{ PAP.adh, }\AttributeTok{parallel =} \ConstantTok{TRUE}\NormalTok{, }\AttributeTok{seed =} \DecValTok{1}\NormalTok{)}
\end{Highlighting}
\end{Shaded}

\hypertarget{evaluation}{%
\subsection{Evaluation}\label{evaluation}}

\hypertarget{assessing-a-cluster-result}{%
\subsubsection{Assessing a cluster
result}\label{assessing-a-cluster-result}}

A cluster result is useful only when it describes the data adequately.
There are various aspects on which the cluster result can be evaluated,
depending on the method and analysis domain:

\begin{itemize}
\tightlist
\item
  The identified solution may not be reliable when the method estimation
  procedure did not converge. Convergence can be checked via the
  \texttt{converged()} function.
\item
  The cluster solution may comprise empty clusters or clusters with a
  negligible proportion of trajectories. In such case, re-estimating the
  method may yield a better solution. Alternatively, one should consider
  fitting the method with a lower number of clusters.
\item
  The cluster trajectories may be assessed visually to determine whether
  the identified patterns are sufficiently distinct.
\item
  The prediction error may help to determine to which degree
  trajectories are represented by one of the clusters.
\end{itemize}

As shown in the previous section, the summary of an \texttt{lcModel}
object shows the method arguments values, cluster sizes, cluster
proportions, cluster names, and the standardized residuals. By default,
the residuals are computed from the difference between the reference
values and the predictions outputted by \texttt{fitted()}, conditional
on the most likely trajectory assignments. For methods that do not
provide trajectory-specific predictions, the fitted values are
determined from the cluster trajectories.

\begin{figure}

{\centering \includegraphics{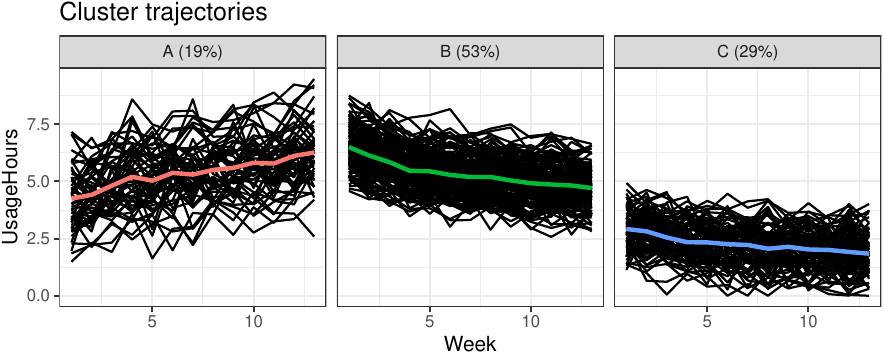} 

}

\caption{The cluster trajectories of the three-cluster solution identified by LMKM, created using `plot(lmkm2)`.}\label{fig:trends}
\end{figure}

The cluster trajectories can be obtained using the
\texttt{clusterTrajectories()} function, returning a
\texttt{data.frame}. The cluster trajectories can be plotted via
\texttt{plot()} or \texttt{plotClusterTrajectories()}. The three-cluster
LMKM solution is visualized in Figure \ref{fig:trends}. For parametric
cluster methods, a more concise representation of the model can be
obtained from the model coefficients, using \texttt{coef()}.

\begin{Shaded}
\begin{Highlighting}[]
\FunctionTok{plot}\NormalTok{(lmkm3, }\AttributeTok{size =} \DecValTok{1}\NormalTok{)}
\end{Highlighting}
\end{Shaded}

Assigning descriptive names to the clusters can help to increase the
readability of the cluster result, which is especially useful for
solutions with many clusters. The \texttt{clusterNames()} function can
be used to retrieve or change the cluster names.

\begin{Shaded}
\begin{Highlighting}[]
\FunctionTok{clusterNames}\NormalTok{(lmkm3) }\OtherTok{\textless{}{-}} \FunctionTok{c}\NormalTok{(}\StringTok{"Struggling"}\NormalTok{, }\StringTok{"Increasing"}\NormalTok{, }\StringTok{"Decreasing"}\NormalTok{)}
\end{Highlighting}
\end{Shaded}

The most likely cluster for each of the trajectories is obtained using
the \texttt{trajectoryAssignments()} function, which outputs a
\texttt{factor} with the cluster names as its levels. For soft-cluster
representations, the cluster assignments are determined by the cluster
with the highest probability, based on the posterior probability matrix.
An alternative approach can be specified through the \texttt{strategy}
argument. For example, the \texttt{which.weight()} function assigns a
random cluster weighted by the proportions. The \texttt{which.is.max()}
function from the \texttt{nnet}
package\footnote{\url{https://CRAN.R-project.org/package=nnet}}
\citep{venables2002modern} returns the most likely cluster, breaking
ties at random.

The posterior probability matrix can be obtained from the
\texttt{postprob()}
function\footnote{For methods that only support modal assignment, the posterior probability matrix only comprises 0 and 1.}.
For probabilistic methods, it can be used to gauge the cluster
separation, i.e., the certainty of assignment. The posterior probability
is also important in the post-hoc analysis for accounting for the
uncertainty in cluster assignment.

When it comes to longitudinal representation, the minimum functionality
that is available for all \texttt{lcModel} objects is the prediction of
the cluster trajectories at the given moments in time. The prediction
has been implemented for underlying packages that lack this
functionality. For non-parametric methods such as KmL or LLPA, linear
interpolation is used when time points are requested which are not
represented by the cluster centers.The available functionality differs
between methods.

All \texttt{lcModel} objects support the standard model functions from
the standard \texttt{stats} package, including \texttt{fitted()},
\texttt{residuals()}, and \texttt{predict()}. These functions are
primarily of interest for methods that have a notion of a group or
individual trajectory prediction error, such as for the regression-based
approaches like GBTM and GMM. The \texttt{fitted()} function returns the
expected values for the response variable for the data on which the
model was estimated. By default, only the values for the most likely
cluster are given. However, for \texttt{clusters\ =\ NULL}, a matrix of
predictions is outputted, where each column represents the predictions
of the respective cluster.

The \texttt{predict()} function computes trajectory- and
cluster-specific predictions for the given input data.

\begin{Shaded}
\begin{Highlighting}[]
\FunctionTok{predict}\NormalTok{(lmkm3, }\AttributeTok{newdata =} \FunctionTok{data.frame}\NormalTok{(}\AttributeTok{Week =} \FunctionTok{c}\NormalTok{(}\DecValTok{1}\NormalTok{, }\DecValTok{10}\NormalTok{), }\AttributeTok{Cluster =} \StringTok{"Decreasing"}\NormalTok{))}
\end{Highlighting}
\end{Shaded}

\begin{verbatim}
##        Fit
## 1 2.919423
## 2 2.024865
\end{verbatim}

The \texttt{predictPostprob()} and \texttt{predictAssignments()}
functions compute the posterior probability and cluster membership for
new trajectories, respectively. As this is not a common use case for
cluster methods, most of the underlying packages do not provide this
functionality. For demonstration purposes, we have implemented the
functionality for the \texttt{lcModelKML} class.

Using the metric interface defined in Section
\protect\hyperlink{sec:methods}{2}, we can compute a variety of internal
metrics through the \texttt{metric()} function:

\begin{Shaded}
\begin{Highlighting}[]
\FunctionTok{metric}\NormalTok{(lmkm3, }\FunctionTok{c}\NormalTok{(}\StringTok{"MAE"}\NormalTok{, }\StringTok{"RMSE"}\NormalTok{, }\StringTok{"Dunn"}\NormalTok{, }\StringTok{"ASW"}\NormalTok{))}
\end{Highlighting}
\end{Shaded}

\begin{verbatim}
##        MAE       RMSE       Dunn        ASW 
## 0.74262252 0.94094913 0.09173111 0.35605235
\end{verbatim}

With a regression-based approach, another aspect that is worthwhile to
assess are the residuals of the predicted values. This can be
investigated, for example, through a visual inspection using a
quantile-quantile (Q-Q) plot, available via the \texttt{qqPlot()}
function, to assess whether the prediction errors approximately follow a
normal distribution.

\hypertarget{identifying-the-number-of-clusters-1}{%
\subsubsection{Identifying the number of
clusters}\label{identifying-the-number-of-clusters-1}}

Using one or more internal metrics of interest, we can assess how the
data representation of a method improves or worsens for an increasing
number of clusters. In this case study, we will use the Dunn index as
the primary metric for the choice of the number of clusters.

The change in metrics for an increasing number of clusters can be
visualized via the \texttt{plotMetric()} function, and can help to
determine the preferred solution. For brevity, we will only provide a
detailed view for the KmL method. We plot the Dunn index, WMAE, and
estimation time (in seconds) for the six KmL solutions as follows:

\begin{Shaded}
\begin{Highlighting}[]
\FunctionTok{plotMetric}\NormalTok{(kmlList, }\FunctionTok{c}\NormalTok{(}\StringTok{"Dunn"}\NormalTok{, }\StringTok{"WMAE"}\NormalTok{, }\StringTok{"estimationTime"}\NormalTok{))}
\end{Highlighting}
\end{Shaded}

\begin{figure}

{\centering \includegraphics{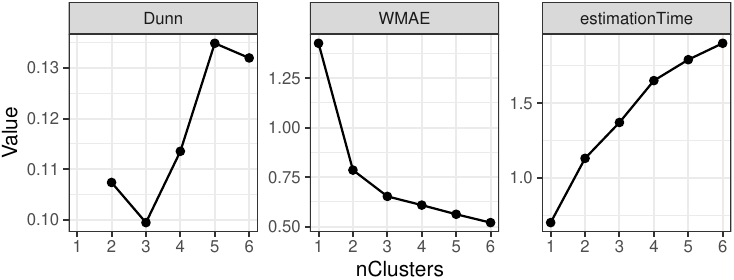} 

}

\caption{The Dunn index (higher is better), and WMAE (lower is better) metrics for each of the KmL solutions from 1 to 6 clusters}\label{fig:numclusmetrics}
\end{figure}

The resulting plot is shown in Figure \ref{fig:numclusmetrics}. The Dunn
index and WMAE show a rather convincing improvement for an increasing
number of
clusters\footnote{The Dunn index is not defined for a one-cluster solution.}.

Moreover, we observe that the estimation time increases with the number
of clusters. This can be a practical consideration when deciding on the
preferred method to use. For much larger datasets, it may be useful to
conduct a preliminary analysis on a subset of the data for possibly
ruling out methods which are too computationally intensive in relation
to the results.

We can obtain the metric values for each of the models by calling the
\texttt{metric()} function.

\begin{Shaded}
\begin{Highlighting}[]
\FunctionTok{metric}\NormalTok{(kmlList, }\FunctionTok{c}\NormalTok{(}\StringTok{"Dunn"}\NormalTok{, }\StringTok{"WMAE"}\NormalTok{, }\StringTok{"estimationTime"}\NormalTok{))}
\end{Highlighting}
\end{Shaded}

\begin{verbatim}
##         Dunn      WMAE estimationTime
## 1         NA 1.4261264           0.70
## 2 0.10737225 0.7850566           1.13
## 3 0.09944419 0.6523208           1.37
## 4 0.11353357 0.6081128           1.65
## 5 0.13487175 0.5619086           1.79
## 6 0.13196444 0.5197172           1.90
\end{verbatim}

As the preferred solution corresponds to the highest Dunn index, we can
obtain the respective model by calling the \texttt{max()} function on
the \texttt{lcModels} list object.

\begin{Shaded}
\begin{Highlighting}[]
\NormalTok{kmlBest }\OtherTok{\textless{}{-}} \FunctionTok{max}\NormalTok{(kmlList, }\StringTok{"Dunn"}\NormalTok{)}
\end{Highlighting}
\end{Shaded}

Alternatively, we can select the preferred model using the
\texttt{subset()} function. By specifying the \texttt{drop\ =\ TRUE},
the \texttt{lcModel} object is returned instead of a \texttt{lcModels}
object.

\begin{Shaded}
\begin{Highlighting}[]
\NormalTok{kmlBest }\OtherTok{\textless{}{-}} \FunctionTok{subset}\NormalTok{(kmlList, nClusters }\SpecialCharTok{==} \DecValTok{5}\NormalTok{, }\AttributeTok{drop =} \ConstantTok{TRUE}\NormalTok{)}
\end{Highlighting}
\end{Shaded}

The identification of the number of clusters is a form of model
selection. The same approach can therefore be used for identifying the
best cluster representation, e.g., evaluating different formulas for a
parametric model, or selecting a different method initialization
strategy.

\hypertarget{comparing-methods}{%
\subsubsection{Comparing methods}\label{comparing-methods}}

The optimal number of clusters according to the internal metric can be
different for other methods or specifications thereof. Depending on the
cluster representation, some methods may require fewer or more clusters
to represent the heterogeneity to the same degree. By concatenating the
lists of fitted methods, we can create a metric plot that is grouped by
the type of method as follows:

\begin{Shaded}
\begin{Highlighting}[]
\NormalTok{allList }\OtherTok{\textless{}{-}} \FunctionTok{lcModels}\NormalTok{(lmkmList, kmlList, dtwList, gbtmList, gmmList)}
\FunctionTok{plotMetric}\NormalTok{(allList, }\AttributeTok{name =} \FunctionTok{c}\NormalTok{(}\StringTok{"Dunn"}\NormalTok{, }\StringTok{"WMAE"}\NormalTok{, }\StringTok{"BIC"}\NormalTok{, }\StringTok{"estimationTime"}\NormalTok{), }\AttributeTok{group =} \StringTok{\textquotesingle{}.method\textquotesingle{}}\NormalTok{)}
\end{Highlighting}
\end{Shaded}

\begin{figure}

{\centering \includegraphics{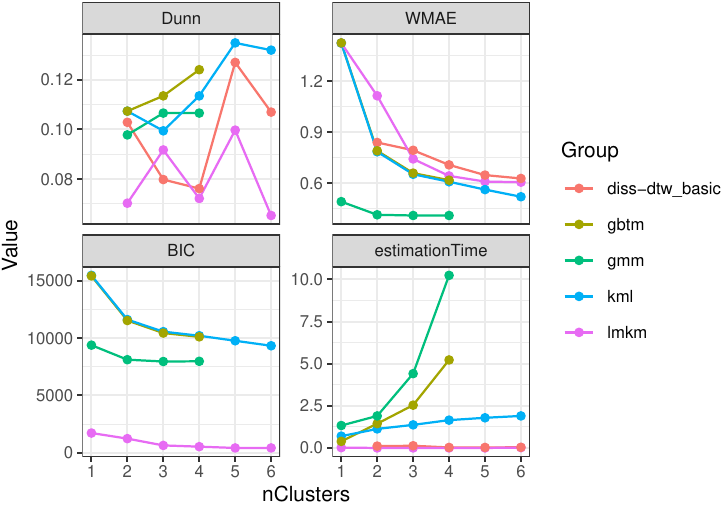} 

}

\caption{The Dunn index (higher is better), WMAE (lower is better) and BIC (relatively lower is better) for each of the methods and number of clusters}\label{fig:numclus}
\end{figure}

The WMAE and BIC between GBTM and KmL are almost exactly the same,
possibly indicating that the methods find a similar solution. If the
solutions are found to be practically identical, then one could actually
prefer KmL due to its considerably favorable computational scaling with
the number of clusters.

We explore the best solution of each method further to better understand
how the cluster representations differ between the methods. We can
select the preferred \texttt{lcModel} object corresponding to the
selected number of clusters for each of the methods using the
\texttt{subset()} function.

\begin{Shaded}
\begin{Highlighting}[]
\NormalTok{kmlBest  }\OtherTok{\textless{}{-}} \FunctionTok{subset}\NormalTok{(kmlList,  nClusters }\SpecialCharTok{==} \DecValTok{5}\NormalTok{, }\AttributeTok{drop =} \ConstantTok{TRUE}\NormalTok{)}
\NormalTok{dtwBest  }\OtherTok{\textless{}{-}} \FunctionTok{subset}\NormalTok{(dtwList,  nClusters }\SpecialCharTok{==} \DecValTok{5}\NormalTok{, }\AttributeTok{drop =} \ConstantTok{TRUE}\NormalTok{)}
\NormalTok{gbtmBest }\OtherTok{\textless{}{-}} \FunctionTok{subset}\NormalTok{(lmkmList, nClusters }\SpecialCharTok{==} \DecValTok{4}\NormalTok{, }\AttributeTok{drop =} \ConstantTok{TRUE}\NormalTok{)}
\NormalTok{lmkmBest }\OtherTok{\textless{}{-}} \FunctionTok{subset}\NormalTok{(lmkmList, nClusters }\SpecialCharTok{==} \DecValTok{3}\NormalTok{, }\AttributeTok{drop =} \ConstantTok{TRUE}\NormalTok{)}
\NormalTok{gmmBest  }\OtherTok{\textless{}{-}} \FunctionTok{subset}\NormalTok{(gmmList,  nClusters }\SpecialCharTok{==} \DecValTok{3}\NormalTok{, }\AttributeTok{drop =} \ConstantTok{TRUE}\NormalTok{)}
\end{Highlighting}
\end{Shaded}

We can then assess the pairwise ARI between each method using the
\texttt{externalMetric()} function. Calling this function on a
\texttt{lcModels} list returns a \texttt{dist} object representing a
distance matrix. We therefore create a list of the best \texttt{lcModel}
for each method, by which we can then determine the pairwise ARI as
follows:

\begin{Shaded}
\begin{Highlighting}[]
\NormalTok{bestList }\OtherTok{\textless{}{-}} \FunctionTok{lcModels}\NormalTok{(}\AttributeTok{KmL =}\NormalTok{ kmlBest, }\AttributeTok{DTW =}\NormalTok{ dtwBest, }
  \AttributeTok{LMKM =}\NormalTok{ lmkmBest, }\AttributeTok{GBTM =}\NormalTok{ gbtmBest, }\AttributeTok{GMM =}\NormalTok{ gmmBest)}
\FunctionTok{externalMetric}\NormalTok{(bestList, }\AttributeTok{name =} \StringTok{"adjustedRand"}\NormalTok{) }\SpecialCharTok{|\textgreater{}} \FunctionTok{signif}\NormalTok{(}\DecValTok{2}\NormalTok{)}
\end{Highlighting}
\end{Shaded}

\begin{verbatim}
##       KmL  DTW LMKM GBTM
## DTW  0.39               
## LMKM 0.48 0.40          
## GBTM 0.65 0.31 0.67     
## GMM  0.49 0.40 0.99 0.68
\end{verbatim}

With all pairwise ARI being at least 0.31, all methods demonstrate some
degree of similarity between each other. In particular, the very high
ARI of approximately 0.99 between GMM and LMKM implies that the methods
grouped the trajectories in a highly similar way.

Secondly, we compare the similarity of the cluster trajectories between
the methods using the WMMAE. The easiest way to compare methods is to
compare the cluster trajectories visually. However, this approach is
only practical on smaller datasets or solutions with few clusters. As a
more scalable alternative, we can use external metrics to measure the
pairwise similarity between the cluster trajectories of the methods.

\begin{Shaded}
\begin{Highlighting}[]
\FunctionTok{externalMetric}\NormalTok{(bestList, }\AttributeTok{name =} \StringTok{"WMMAE"}\NormalTok{) }\SpecialCharTok{|\textgreater{}} \FunctionTok{signif}\NormalTok{(}\DecValTok{2}\NormalTok{)}
\end{Highlighting}
\end{Shaded}

\begin{verbatim}
##        KmL   DTW  LMKM  GBTM
## DTW  0.100                  
## LMKM 0.140 0.130            
## GBTM 0.068 0.130 0.091      
## GMM  0.130 0.130 0.036 0.099
\end{verbatim}

The mean absolute error of 0.091 between the cluster trajectories of
GBTM and LMKM is negligible compared to the residual error estimated by
GBTM (\(\textrm{SD} = 0.8\)), which indicates that both methods have
identified practically the same cluster trajectories. The same applies
to GMM and LMKM.

\hypertarget{cluster-validation}{%
\subsection{Cluster validation}\label{cluster-validation}}

Assessing the stability and reproducibility of a cluster method can help
to determine whether the identified cluster solution generalizes beyond
the data that was used to estimate the method. This is especially
relevant for more complex cluster methods involving many parameters,
which may not generalize well to new data. This primarily pertains to
the number of clusters the method is estimated for, as the number of
parameters increases linearly with the number of clusters. Even
relatively simple methods can overfit the data when the representation
comprises too many clusters in relation to the sample size.

\hypertarget{cluster-stability-using-repeated-estimation}{%
\subsubsection{Cluster stability using repeated
estimation}\label{cluster-stability-using-repeated-estimation}}

Many of the cluster methods can yield a slightly different solution
during each run, depending on the starting conditions. In such cases, by
doing a repeated estimation, we can gauge the stability, i.e.,
consistency, of the method. Comparing repeated estimation results is
also useful for selecting the best solution for a given method.

Repeated estimation can be done via the \texttt{latrendRep()} function,
where the number of repetitions is specified via the \texttt{.rep}
argument. Similar to \texttt{latrend()}, the method arguments can be
updated within the function. The function returns a \texttt{lcModels}
object, comprising a \texttt{list} of \texttt{lcModel} objects. Here, we
only use five repeated estimations to limit the computation time. In
practice, a higher number such as 10 or 25 is advisable, depending on
the magnitude of instability.

\begin{Shaded}
\begin{Highlighting}[]
\NormalTok{kmlRepList }\OtherTok{\textless{}{-}} \FunctionTok{latrendRep}\NormalTok{(kmlMethod, }\AttributeTok{data =}\NormalTok{ PAP.adh, }
  \AttributeTok{nClusters =} \DecValTok{5}\NormalTok{, }\AttributeTok{.rep =} \DecValTok{5}\NormalTok{, }\AttributeTok{.parallel =} \ConstantTok{TRUE}\NormalTok{)}
\FunctionTok{summary}\NormalTok{(}\FunctionTok{metric}\NormalTok{(kmlRepList, }\FunctionTok{c}\NormalTok{(}\StringTok{"Dunn"}\NormalTok{, }\StringTok{"WMAE"}\NormalTok{)))}
\end{Highlighting}
\end{Shaded}

\begin{verbatim}
##       Dunn             WMAE       
##  Min.   :0.1047   Min.   :0.5599  
##  1st Qu.:0.1349   1st Qu.:0.5610  
##  Median :0.1349   Median :0.5617  
##  Mean   :0.1288   Mean   :0.5618  
##  3rd Qu.:0.1349   3rd Qu.:0.5619  
##  Max.   :0.1349   Max.   :0.5647
\end{verbatim}

The result suggests that the solutions found by KmL for the given number
of clusters has some degree of variability, which could indicate that
different solutions are found for the same number of clusters.

\hypertarget{cluster-stability-using-bootstrap-sampling}{%
\subsubsection{Cluster stability using bootstrap
sampling}\label{cluster-stability-using-bootstrap-sampling}}

Instead of assessing the cluster stability across repeated estimation on
the same dataset, we can obtain a more generalizable estimate of the
cluster stability by measuring the cluster stability across different
datasets. Bootstrap sampling, also referred to as bootstrapping,
involves the repeated estimation on simulated datasets generated from
the original dataset. It is primarily used for assessing the stability
of a method, as measured by one or more internal metrics. Here, complete
trajectories are selected at random with replacement from the dataset to
generate a new dataset of equal size. Each simulated dataset, referred
to as a bootstrap sample in this context, will yield a slightly
different solution. This variability between samples can provide an
indication of the stability of the cluster method on the overall dataset
\citep{hennig2007clusterwise}. Since the repeated estimation is done on
new datasets that only partially
overlap\footnote{In addition to the challenge of the cluster representations being in a different order between runs, also referred to as label switching.},
this restricts the available external metrics to only those that can
compare between different datasets, e.g., the WMMAE.

The \texttt{latrendBoot()} function applies bootstrapping to the given
method specification. The \texttt{samples} argument determines the
number of times the data is resampled, and a model is estimated. Setting
the \texttt{seed} argument ensures that the same sequence of bootstrap
samples is generated when redoing the bootstrapping procedure. The
output is a \texttt{lcModels} list containing the model for each sample.
The estimated methods each have a different \texttt{call} for the
\texttt{data} argument such that the original bootstrap training sample
can be recreated as needed. This avoids the need for models to store the
training data. As an example, we compute 20 bootstrap
samples\footnote{In practice, a much greater number of bootstrap samples is recommended (at least 100).}
(i.e., repeated fits) in parallel as follows:

\begin{Shaded}
\begin{Highlighting}[]
\NormalTok{kmlMethodBest }\OtherTok{\textless{}{-}} \FunctionTok{update}\NormalTok{(kmlMethod, }\AttributeTok{nClusters =} \DecValTok{5}\NormalTok{)}
\NormalTok{kmlBootModels }\OtherTok{\textless{}{-}} \FunctionTok{latrendBoot}\NormalTok{(kmlMethodBest, }\AttributeTok{data =}\NormalTok{ PAP.adh, }
  \AttributeTok{samples =} \DecValTok{10}\NormalTok{, }\AttributeTok{seed =} \DecValTok{1}\NormalTok{, }\AttributeTok{parallel =} \ConstantTok{TRUE}\NormalTok{)}
\FunctionTok{head}\NormalTok{(kmlBootModels, }\AttributeTok{n =} \DecValTok{3}\NormalTok{)}
\end{Highlighting}
\end{Shaded}

\begin{verbatim}
## List of 3 lcModels with
##   .name .method                                        data       seed
## 1     1     kml  bootSample(PAP.adh, "Patient", 762473831L) 1062140483
## 2     2     kml bootSample(PAP.adh, "Patient", 1762587819L)  185557490
## 3     3     kml bootSample(PAP.adh, "Patient", 1463113723L)  934902099
\end{verbatim}

We can now assess the stability of the solutions across the models in
terms of metrics of interest. Here, we assess the mean convergence rate,
and the quantiles of the WMAE and Dunn metrics.

\begin{Shaded}
\begin{Highlighting}[]
\NormalTok{bootMetrics }\OtherTok{\textless{}{-}} \FunctionTok{metric}\NormalTok{(kmlBootModels, }\FunctionTok{c}\NormalTok{(}\StringTok{"converged"}\NormalTok{, }\StringTok{"Dunn"}\NormalTok{, }\StringTok{"WMAE"}\NormalTok{))}
\FunctionTok{mean}\NormalTok{(bootMetrics}\SpecialCharTok{$}\NormalTok{converged)}
\end{Highlighting}
\end{Shaded}

\begin{verbatim}
## [1] 1
\end{verbatim}

\begin{Shaded}
\begin{Highlighting}[]
\FunctionTok{summary}\NormalTok{(bootMetrics}\SpecialCharTok{$}\NormalTok{Dunn)}
\end{Highlighting}
\end{Shaded}

\begin{verbatim}
##    Min. 1st Qu.  Median    Mean 3rd Qu.    Max. 
##  0.1351  0.1477  0.1506  0.1570  0.1688  0.1852
\end{verbatim}

\begin{Shaded}
\begin{Highlighting}[]
\FunctionTok{summary}\NormalTok{(bootMetrics}\SpecialCharTok{$}\NormalTok{WMAE)}
\end{Highlighting}
\end{Shaded}

\begin{verbatim}
##    Min. 1st Qu.  Median    Mean 3rd Qu.    Max. 
##  0.5289  0.5490  0.5553  0.5534  0.5587  0.5736
\end{verbatim}

As can be seen from the output, there is quite some variability between
the estimated solutions across bootstrap samples. This suggests that we
should consider estimation with repeated random starts to identify a
better and more stable solution.

Lastly, we can compute a similarity matrix for an external metric of
interest, containing the pairwise similarity for each model pair.

\begin{Shaded}
\begin{Highlighting}[]
\NormalTok{wmmaeDist }\OtherTok{\textless{}{-}} \FunctionTok{externalMetric}\NormalTok{(kmlBootModels[}\DecValTok{1}\SpecialCharTok{:}\DecValTok{10}\NormalTok{], }\AttributeTok{name =} \StringTok{"WMMAE"}\NormalTok{)}
\FunctionTok{summary}\NormalTok{(wmmaeDist)}
\end{Highlighting}
\end{Shaded}

\begin{verbatim}
##    Min. 1st Qu.  Median    Mean 3rd Qu.    Max. 
## 0.01392 0.06029 0.07294 0.06841 0.07934 0.11060
\end{verbatim}

Showing that there is only a small degree of discrepancy in the cluster
trajectories between bootstrap samples.

\hypertarget{comparison-to-ground-truth}{%
\subsubsection{Comparison to ground
truth}\label{comparison-to-ground-truth}}

We now consider the case where a method is evaluated in a simulation
study. In such a study, the ground truth is known, and we can directly
evaluate whether the trajectories are clustered correctly. A useful and
intuitive measure is the split-join distance
\citep{dongen2000performance}, which is an edit distance that measures
the number of trajectory reassignments that are needed to go from one
partitioning to another. In case of a ground truth, we are only
interested in the edit distance from the reference
partitioning\footnote{In the one-way edit distance, a solution that has more clusters than the reference can still obtain an edit distance of zero if the extra clusters are a subset of the cluster of the reference}.

We can obtain the vector of trajectory cluster membership of the
\texttt{PAP.adh} from the \texttt{Group} column by selecting the first
cluster name of each trajectory, since the cluster membership is stable
over time. We then create a \texttt{lcModelPartition} from the computed
membership vector. By default, the \texttt{lcModelPartition} generates
the cluster representations from the means of the trajectories assigned
to the respective cluster.

\begin{Shaded}
\begin{Highlighting}[]
\NormalTok{refAssignments }\OtherTok{\textless{}{-}} \FunctionTok{aggregate}\NormalTok{(Group }\SpecialCharTok{\textasciitilde{}}\NormalTok{ Patient, }\AttributeTok{data =}\NormalTok{ PAP.adh, }\AttributeTok{FUN =}\NormalTok{ head, }\AttributeTok{n =}\NormalTok{ 1L)}
\NormalTok{refAssignments}\SpecialCharTok{$}\NormalTok{Cluster }\OtherTok{=}\NormalTok{ refAssignments}\SpecialCharTok{$}\NormalTok{Group}

\NormalTok{refModel }\OtherTok{\textless{}{-}} \FunctionTok{lcModelPartition}\NormalTok{(}\AttributeTok{data =}\NormalTok{ PAP.adh, }
  \AttributeTok{trajectoryAssignments =}\NormalTok{ refAssignments, }\AttributeTok{response =} \StringTok{"UsageHours"}\NormalTok{)}
\NormalTok{refModel}
\end{Highlighting}
\end{Shaded}

\begin{verbatim}
## Longitudinal cluster model using part
## lcMethod specifying "undefined"
## no arguments
## 
## Cluster sizes (K=3):
##     Adherers    Improvers Non-adherers 
##  162 (53.8%)   56 (18.6%)   83 (27.6%) 
## 
## Number of obs: 3913, strata (Patient): 301
## 
## Scaled residuals:
##      Min.   1st Qu.    Median      Mean   3rd Qu.      Max. 
## -3.894748 -0.643670 -0.009533  0.000000  0.634893  3.590377
\end{verbatim}

We can now compare our selected method solutions to the reference
solution using the one-way split-join distance to the reference:

\begin{Shaded}
\begin{Highlighting}[]
\FunctionTok{externalMetric}\NormalTok{(bestList, refModel, }\AttributeTok{name =} \StringTok{"splitJoin.ref"}\NormalTok{, }\AttributeTok{drop =} \ConstantTok{FALSE}\NormalTok{)}
\end{Highlighting}
\end{Shaded}

\begin{verbatim}
##      splitJoin.ref
## KmL             24
## DTW             61
## LMKM             3
## GBTM             1
## GMM              2
\end{verbatim}

This shows that, for the \texttt{PAP.adh} dataset, LMKM, GBTM, and GMM
achieve a nearly perfect recovery of the cluster memberships, but that
GBTM needs more clusters to represent the dataset.

\hypertarget{sec:extension}{%
\section{Implementing new methods}\label{sec:extension}}

One of the main strengths of the framework is the standard way in which
methods are specified, estimated, and evaluated. These aspects make it
easy to compare newly implemented methods with existing ones. Using the
base classes \texttt{lcMethod} and \texttt{lcModel}, new methods can be
implemented with a relatively minimal amount of code, enabling rapid
prototyping. These classes provide basic functionality, from which the
user can extend certain functions as needed by creating a subclass.

\hypertarget{stratification}{%
\subsection{Stratification}\label{stratification}}

The simplest form of clustering is the stratification of the dataset
based on a known factor. This can be the response variable, or any other
measure available for each trajectory. This is useful for case studies
where there is prior knowledge or expert guidance on how the
trajectories should be grouped; either by another factor (e.g., age or
gender), or a characteristic of the trajectory (e.g., the intercept,
slope, average, or variance).

A stratification approach can be specified using the
\texttt{lcMethodStratify()} function, which takes an \texttt{R}
expression as input. The expression is evaluated within the
\texttt{data.frame} at the trajectory level during the method
estimation, so any column present in the data can be used. The
expression should resolve to a number or category, indicating the
stratum for the respective trajectory.

As an example, we stratify the trajectories by thresholding on the mean
hours of usage. This expression returns a \texttt{logical} value which
determines the cluster assignment. For categorizing trajectories into
more than two clusters, the \texttt{cut()} function can be used. The
cluster trajectories are computed by aggregating the trajectories of
each cluster at the respective time points. By default, the average is
computed, but an alternative center function can be specified via the
\texttt{center} argument.

\begin{Shaded}
\begin{Highlighting}[]
\NormalTok{stratMethod }\OtherTok{\textless{}{-}} \FunctionTok{lcMethodStratify}\NormalTok{(}\AttributeTok{response =} \StringTok{"UsageHours"}\NormalTok{, }\AttributeTok{stratify =} \FunctionTok{mean}\NormalTok{(UsageHours) }\SpecialCharTok{\textgreater{}} \DecValTok{4}\NormalTok{)}
\NormalTok{stratModel }\OtherTok{\textless{}{-}} \FunctionTok{latrend}\NormalTok{(stratMethod, }\AttributeTok{data =}\NormalTok{ PAP.adh)}
\FunctionTok{clusterProportions}\NormalTok{(stratModel)}
\end{Highlighting}
\end{Shaded}

\begin{verbatim}
##         A         B 
## 0.3156146 0.6843854
\end{verbatim}

\hypertarget{feature-based-clustering-1}{%
\subsection{Feature-based clustering}\label{feature-based-clustering-1}}

Feature-based clustering is a flexible and fast approach to clustering
longitudinal data, with an essentially limitless choice of trajectory
representations. The framework includes a generic feature-based
clustering class named \texttt{lcMethodFeature} for quickly implementing
this approach.

A \texttt{lcMethodFeature} specification requires two functions: A
representation function outputting the trajectory representation
\texttt{matrix}, and a cluster function that applies a cluster algorithm
to the matrix, returning an \texttt{lcModel} object.

To illustrate the method, we represent each trajectory using a linear
model, and we cluster the model coefficients using \(k\)-means. In the
representation step, \texttt{lm()} is applied to each trajectory, and
the model coefficients are combined into a \texttt{matrix} with the
trajectory-specific coefficients on each row. We parameterize the
\texttt{lcMethod} implementation by obtaining the model formula from
\texttt{method\$formula}. During the method specification, the user
therefore needs to define the \texttt{formula} argument. The
representation function is as follows:

\begin{Shaded}
\begin{Highlighting}[]
\NormalTok{repStep }\OtherTok{\textless{}{-}} \ControlFlowTok{function}\NormalTok{(method, data, verbose) \{}
\NormalTok{  repTraj }\OtherTok{\textless{}{-}} \ControlFlowTok{function}\NormalTok{(trajData) \{}
\NormalTok{    lm.rep }\OtherTok{\textless{}{-}} \FunctionTok{lm}\NormalTok{(method}\SpecialCharTok{$}\NormalTok{formula, }\AttributeTok{data =}\NormalTok{ trajData)}
    \FunctionTok{coef}\NormalTok{(lm.rep)}
\NormalTok{  \}}
\NormalTok{  dt }\OtherTok{\textless{}{-}} \FunctionTok{as.data.table}\NormalTok{(data)}
\NormalTok{  coefData }\OtherTok{\textless{}{-}}\NormalTok{ dt[, }\FunctionTok{as.list}\NormalTok{(}\FunctionTok{repTraj}\NormalTok{(.SD)), keyby }\OtherTok{=} \FunctionTok{c}\NormalTok{(method}\SpecialCharTok{$}\NormalTok{id)]}
\NormalTok{  coefMat }\OtherTok{\textless{}{-}} \FunctionTok{as.matrix}\NormalTok{(}\FunctionTok{subset}\NormalTok{(coefData, }\AttributeTok{select =} \SpecialCharTok{{-}}\DecValTok{1}\NormalTok{))}
  \FunctionTok{rownames}\NormalTok{(coefMat) }\OtherTok{\textless{}{-}}\NormalTok{ coefData[[method}\SpecialCharTok{$}\NormalTok{id]]}
\NormalTok{  coefMat}
\NormalTok{\}}
\end{Highlighting}
\end{Shaded}

We implement the cluster step to return a \texttt{lcModelPartition}
object based on the cluster assignments outputted by \texttt{kmeans()}.
We have parameterized the function by obtaining the number of clusters
for \(k\)-means from the \texttt{nClusters} model argument. The cluster
function is as follows:

\begin{Shaded}
\begin{Highlighting}[]
\NormalTok{clusStep }\OtherTok{\textless{}{-}} \ControlFlowTok{function}\NormalTok{(method, data, repMat, envir, verbose) \{}
\NormalTok{  km }\OtherTok{\textless{}{-}} \FunctionTok{kmeans}\NormalTok{(repMat, }\AttributeTok{centers =}\NormalTok{ method}\SpecialCharTok{$}\NormalTok{nClusters)}
  \FunctionTok{lcModelPartition}\NormalTok{(}\AttributeTok{response =} \FunctionTok{responseVariable}\NormalTok{(method), }\AttributeTok{method =}\NormalTok{ method,}
    \AttributeTok{data =}\NormalTok{ data, }\AttributeTok{trajectoryAssignments =}\NormalTok{ km}\SpecialCharTok{$}\NormalTok{cluster, }\AttributeTok{center =}\NormalTok{ mean)}
\NormalTok{\}}
\end{Highlighting}
\end{Shaded}

We can now specify and estimate the feature-based method, including the
additionally required arguments. Comparing the estimated model to the
preferred KmL model, we see that the solutions have a relatively high
degree of overlap.

\begin{Shaded}
\begin{Highlighting}[]
\NormalTok{tsMethod }\OtherTok{\textless{}{-}} \FunctionTok{lcMethodFeature}\NormalTok{(}\AttributeTok{response =} \StringTok{"UsageHours"}\NormalTok{, }\AttributeTok{formula =}\NormalTok{ UsageHours }\SpecialCharTok{\textasciitilde{}}\NormalTok{ Week, }
  \AttributeTok{representationStep =}\NormalTok{ repStep, }\AttributeTok{clusterStep =}\NormalTok{ clusStep)}
\NormalTok{tsModel }\OtherTok{\textless{}{-}} \FunctionTok{latrend}\NormalTok{(tsMethod, }\AttributeTok{data =}\NormalTok{ PAP.adh, }\AttributeTok{nClusters =} \DecValTok{5}\NormalTok{)}
\FunctionTok{externalMetric}\NormalTok{(tsModel, kmlBest, }\StringTok{"adjustedRand"}\NormalTok{)}
\end{Highlighting}
\end{Shaded}

\begin{verbatim}
## adjustedRand 
##    0.4487578
\end{verbatim}

\begin{Shaded}
\begin{Highlighting}[]
\FunctionTok{externalMetric}\NormalTok{(tsModel, kmlBest, }\StringTok{"WMMAE"}\NormalTok{)}
\end{Highlighting}
\end{Shaded}

\begin{verbatim}
##      WMMAE 
## 0.08127578
\end{verbatim}

\hypertarget{sec:custommethod}{%
\subsection{Implementing a method}\label{sec:custommethod}}

The framework is designed to support the implementation of new methods,
so that users can extend or implement new methods to address their use
case. In this section, we describe the high-level steps that are
involved in adding support for a method to the framework. Considering
the number of lines of code for even a relatively simple cluster method,
we do not cover a complete example here. Instead, we only outline the
typical set of functions (\texttt{fit()},
\texttt{getArgumentDefaults()}, \texttt{getName()},
\texttt{getShortName()}) that need to be implemented, together with any
relevant input and output assumptions of these functions. For complete
examples, see the \texttt{lcMethod}-interface implementations based on
external packages, e.g., \texttt{lcMethodKML} or
\texttt{lcMethodLcmmGMM}. A step-by-step example of implementing a
statistical method in the framework can be found in the vignette
included with the package, which can be viewed by running
\texttt{vignette("implement",\ package\ =\ \ "latrend")}.

The estimation process of a method is divided into six steps, involving
the processing of the method arguments, preparing and validating the
data, and fitting the specified method. All steps except for
\texttt{fit()} are optional.

\begin{enumerate}
\def\labelenumi{\arabic{enumi}.}
\tightlist
\item
  The \texttt{prepareData()} function transforms the training data into
  the required format for the internal method estimation code. By
  default, data is provided in long format in a \texttt{data.frame}. For
  most implementations, no transformation is therefore needed. Cluster
  methods for repeated-measures data typically require data to be
  transformed to \texttt{matrix} format, however.
\item
  The \texttt{compose()} function evaluates the method arguments and
  returns an updated \texttt{lcMethod} object with the evaluated method
  arguments. The function can also be used for modifying or even
  replacing the original \texttt{lcMethod} object for the remainder of
  the estimation process. This is useful when a method is a special case
  of a more general method and intends to conceal derivative or
  redundant arguments from the base class.
\item
  The \texttt{validate()} function enables evaluated method arguments to
  be checked against the input data. This can be used, for example, for
  checking whether the data contains the covariates specified in the
  method formula, or whether an argument has a valid value. For
  implementations which wrap an underlying package function, this
  validation is usually not needed as the underlying package already
  performs validation of the input.
\item
  The \texttt{preFit()} function is intended for processing any
  arguments prior to fitting. In order for these results to be
  persistent, they should be returned in an \texttt{environment} object,
  which will be passed as an input to the \texttt{fit()} function.
\item
  The \texttt{fit()} function is where the internal method is estimated
  for the given specification to obtain the cluster result. This
  function is also responsible for creating the corresponding
  \texttt{lcModel} object. The running time of this function is used to
  determine the method estimation time.
\item
  The \texttt{postFit()} function takes the outputted \texttt{lcModel}
  from \texttt{fit()} as input, enabling post-processing to be done.
  This is used, for example, for computing derivative statistics, or for
  reducing the memory footprint by stripping redundant data fields from
  the internal model representation. Preferably, this function is
  implemented such that it can be called repeatedly, allowing for
  updates to fitted methods without requiring re-estimation.
\end{enumerate}

The implementation of a method requires defining a new \texttt{lcMethod}
class, which we will name \texttt{lcMethodExample} here. Usually, a new
\texttt{lcModel} class needs to be implemented to handle the result and
representation of the fitted method, which we will name
\texttt{lcModelExample} in the example below. If the new method only
outputs a partitioning, then the \texttt{lcModelPartition} class may be
used instead.

\hypertarget{extending-the-method-class}{%
\subsubsection{Extending the method
class}\label{extending-the-method-class}}

Defining a new method involves creating a subclass of the
\texttt{lcMethod} class, defining its default arguments, its name, and
any logic needed for the fitting procedure.

We start by defining the \texttt{lcMethodExample} class.

\begin{Shaded}
\begin{Highlighting}[]
\FunctionTok{setClass}\NormalTok{(}\StringTok{"lcMethodExample"}\NormalTok{, }\AttributeTok{contains =} \StringTok{"lcMethod"}\NormalTok{)}
\end{Highlighting}
\end{Shaded}

Any method can be specified by instantiating the respective class
through the \texttt{new()} function. It is recommended to rely on the
object initialization mechanism of the base \texttt{lcMethod} class for
this, as it takes care of collecting all arguments and adding default
values for missing arguments. Defining new method arguments in custom
class slots would hinder users from passing specialized or new optional
arguments to the underlying estimation call.

Given that the base class handles the initialization of our
\texttt{lcMethodExample} class, all we need to do is to define the
default argument values in a named list. By adding
\texttt{formals(stats::kmeans)} to the named list, our method will
inherit all arguments from the \texttt{kmeans()} function.

\begin{Shaded}
\begin{Highlighting}[]
\FunctionTok{setMethod}\NormalTok{(}\StringTok{"getArgumentDefaults"}\NormalTok{, }\StringTok{"lcMethodExample"}\NormalTok{, }\ControlFlowTok{function}\NormalTok{(object) \{}
  \FunctionTok{c}\NormalTok{(}
    \FunctionTok{formals}\NormalTok{(stats}\SpecialCharTok{::}\NormalTok{kmeans),}
    \AttributeTok{time =} \FunctionTok{quote}\NormalTok{(}\FunctionTok{getOption}\NormalTok{(}\StringTok{"latrend.time"}\NormalTok{)),}
    \AttributeTok{id =} \FunctionTok{quote}\NormalTok{(}\FunctionTok{getOption}\NormalTok{(}\StringTok{"latrend.id"}\NormalTok{)), }
    \AttributeTok{nClusters =} \DecValTok{2}\NormalTok{,}
    \FunctionTok{callNextMethod}\NormalTok{()}
\NormalTok{  )}
\NormalTok{\})}
\end{Highlighting}
\end{Shaded}

Method arguments can be of any class. However, we recommend that methods
are specified using scalar arguments. This results in a more easily
readable method summary, and greatly simplifies the permutation of
argument options in a simulation study.

For identification purposes, it is recommended to specify a name and an
abbreviated name for the method. This can be done by implementing the
\texttt{getName()} and \texttt{getShortName()} functions, returning the
names as \texttt{character}.

\begin{Shaded}
\begin{Highlighting}[]
\FunctionTok{setMethod}\NormalTok{(}\StringTok{"getName"}\NormalTok{, }\StringTok{"lcMethodExample"}\NormalTok{, }\ControlFlowTok{function}\NormalTok{(object) }\StringTok{"simple example method"}\NormalTok{)}

\FunctionTok{setMethod}\NormalTok{(}\StringTok{"getShortName"}\NormalTok{, }\StringTok{"lcMethodExample"}\NormalTok{, }\ControlFlowTok{function}\NormalTok{(object) }\StringTok{"example"}\NormalTok{)}
\end{Highlighting}
\end{Shaded}

We can now specify our example method by instantiating an object through
the \texttt{new()} function, providing optional arguments as additional
inputs.

\begin{Shaded}
\begin{Highlighting}[]
\FunctionTok{new}\NormalTok{(}\StringTok{"lcMethodExample"}\NormalTok{, }\AttributeTok{nClusters =} \DecValTok{3}\NormalTok{)}
\end{Highlighting}
\end{Shaded}

\begin{verbatim}
## lcMethodExample specifying "simple example method"
##  iter.max:       10
##  nstart:         1
##  algorithm:      c("Hartigan-Wong", "Lloyd", "Forgy", "Ma
##  trace:          FALSE
##  time:           getOption("latrend.time")
##  id:             getOption("latrend.id")
##  nClusters:      3
\end{verbatim}

At the very least, we need to define a \texttt{fit()} function which
uses the \texttt{lcMethodExample} object passed via the \texttt{method}
argument and the data to estimate the specified model. The function
returns a new \texttt{lcModelExample} object based on the internal
model.

\begin{Shaded}
\begin{Highlighting}[]
\FunctionTok{setMethod}\NormalTok{(}\StringTok{"fit"}\NormalTok{, }\StringTok{"lcMethodExample"}\NormalTok{, }
  \ControlFlowTok{function}\NormalTok{(method, data, envir, verbose, ...) \{}
\NormalTok{    fittedRepresentation }\OtherTok{\textless{}{-}}\NormalTok{ CODE\_HERE}
    \FunctionTok{new}\NormalTok{(}\StringTok{"lcModelExample"}\NormalTok{, }\AttributeTok{data =}\NormalTok{ data, }\AttributeTok{model =}\NormalTok{ fittedRepresentation, }
      \AttributeTok{method =}\NormalTok{ method, }\AttributeTok{clusterNames =} \FunctionTok{make.clusterNames}\NormalTok{(method}\SpecialCharTok{$}\NormalTok{nClusters)}
\NormalTok{)\})}
\end{Highlighting}
\end{Shaded}

In case an external estimation function should be called with the
defined method arguments, one can apply \texttt{as.list()} to the
\texttt{lcMethod} object to obtain a named list of argument values. The
external function can then be called using \texttt{do.call()}.

Checking for missing arguments and for the correct argument type or
valid values avoids late and confusing errors during the estimation
process. It is therefore recommended to implement a validation mechanism
of the method specification. This can be done by assigning a validation
function to the class via \texttt{setValidity()} as part of the S4
system, or by implementing \texttt{validate()}. The latter function
allows for easier validation as all arguments are already evaluated, and
the arguments can be validated against the input data.

\hypertarget{extending-the-model-class}{%
\subsubsection{Extending the model
class}\label{extending-the-model-class}}

We begin by defining the \texttt{lcModelExample} class. One can consider
adding slots for representing, for example, the representational
coefficients.

\begin{Shaded}
\begin{Highlighting}[]
\FunctionTok{setClass}\NormalTok{(}\StringTok{"lcModelExample"}\NormalTok{, }\AttributeTok{contains =} \StringTok{"lcModel"}\NormalTok{)}
\end{Highlighting}
\end{Shaded}

The \texttt{postprob()} function is used to determine the cluster
assignments and cluster proportions, so every \texttt{lcModel} subclass
should provide it. In case of hard-cluster models, the posterior
probability consists of zeros and ones.

\begin{Shaded}
\begin{Highlighting}[]
\FunctionTok{setMethod}\NormalTok{(}\StringTok{"postprob"}\NormalTok{, }\StringTok{"lcModelExample"}\NormalTok{, }\ControlFlowTok{function}\NormalTok{(object) \{}
\NormalTok{  ppMatrix }\OtherTok{\textless{}{-}}\NormalTok{ CODE\_HERE}
  \FunctionTok{colnames}\NormalTok{(ppMatrix) }\OtherTok{\textless{}{-}} \FunctionTok{clusterNames}\NormalTok{(object)}
  \FunctionTok{return}\NormalTok{ (ppMatrix)}
\NormalTok{\})}
\end{Highlighting}
\end{Shaded}

The \texttt{predict.lcModel()} function is relatively complex due to the
different types of inputs and outputs it supports. As these cases
generalize across methods, the \texttt{lcModel} class provides a
suitable standard implementation. For implementing new \texttt{lcModel}
classes, it is therefore advisable to implement the
\texttt{predictForCluster()} function instead of \texttt{predict()}, as
it is called by \texttt{predict.lcModel()}. This function should provide
a prediction for each row of the \texttt{data.frame} of the
\texttt{newdata} argument, conditional on the given cluster membership.

\begin{Shaded}
\begin{Highlighting}[]
\FunctionTok{setMethod}\NormalTok{(}\StringTok{"predictForCluster"}\NormalTok{, }\StringTok{"lcModelExample"}\NormalTok{, }
  \ControlFlowTok{function}\NormalTok{(object, newdata, cluster, ...) \{}
\NormalTok{    predData }\OtherTok{\textless{}{-}}\NormalTok{ CODE\_HERE}
    \FunctionTok{return}\NormalTok{ (predData)}
\NormalTok{\})}
\end{Highlighting}
\end{Shaded}

Lastly, implementing the \texttt{predictPostprob()} function enables the
model to predict the posterior probability, and correspondingly the
cluster membership, for new trajectories. The output should be a matrix
matching the number of rows of \texttt{newdata} and indicating the
cluster-specific probabilities in the respective columns.

\begin{Shaded}
\begin{Highlighting}[]
\FunctionTok{setMethod}\NormalTok{(}\StringTok{"predictPostprob"}\NormalTok{, }\StringTok{"lcModelExample"}\NormalTok{, }
  \ControlFlowTok{function}\NormalTok{(object, newdata, ...) \{}
\NormalTok{    ppMat }\OtherTok{\textless{}{-}}\NormalTok{ CODE\_HERE}
    \FunctionTok{colnames}\NormalTok{(ppMat) }\OtherTok{\textless{}{-}} \FunctionTok{clusterNames}\NormalTok{(object)}
    \FunctionTok{return}\NormalTok{ (ppMat)}
\NormalTok{\})}
\end{Highlighting}
\end{Shaded}

It is also possible to override the \texttt{predictAssignments()}
function. However, the default function already uses the output of
\texttt{predictPostprob()}, so overriding it is only of use for
implementing a more extensive or method-specific classification
strategy.

\hypertarget{sec:discussion}{%
\section{Summary and outlook}\label{sec:discussion}}

The \texttt{latrend} package facilitates the standardized yet flexible
exploration of heterogeneity in longitudinal datasets, with a minimal
amount of coding effort. The framework provides functionality for
specifying, estimating, and assessing models for clustering longitudinal
data. The package builds upon the efforts of the \texttt{R} community by
providing an interface to the many methods for clustering longitudinal
data across packages. Perhaps most importantly, the \texttt{latrend}
package makes it easy to compare between any two cluster methods,
enabling users to identify the most suitable method to their use case.
To ensure transparent and reproducible research, all decisions and
settings that are relevant to the analysis should be reported. A useful
checklist for reporting on latent-class trajectory studies is provided
by \citet{van2017grolts}, which is also relevant to longitudinal cluster
analyses in general.

Users can implement new methods within the framework or add support for
other packages, enabling rapid prototyping for the case study at hand.
Additionally, the standard functionality provided by the framework also
reduces the effort needed in implementing a longitudinal cluster model.

We encourage the framework to be used as a first exploratory step in
clustering longitudinal data, after which the identified preferred
method can then be applied directly from the original package, which
typically provides special tools or options not provided by the
framework. To illustrate one such limitation, consider the
initialization or prior specification of a longitudinal cluster model.
This is generally an important aspect of model estimation that can
improve the identified model solution but is challenging to facilitate
in a standardized way.

Although the package allows for the automatic comparison and selection
across methods through various metrics, it is advisable to assess
whether the identified cluster solution is meaningful. It is a useful
practice to consider domain knowledge when evaluating the solution
\citep{nagin2005group}, both in the choice of metrics as well as the
interpretation of the clusters. For example, in some applications, the
change over time is more of interest than the mean level, and vice
versa. Along similar lines, a solution comprising a very small cluster
(i.e., with few subjects) provides little additional descriptive power
of the heterogeneity, unless the presence of outliers is of significant
interest.

The framework is currently focused towards the modeling of a single
continuous response variable, whereas some of the supported cluster
packages already support multitrajectory modeling. The possible support
for multitrajectory modeling has been accounted for in the design of the
software. Similarly, while the single response is required to be
numerical, support could be added for categorical outcomes such as those
used in longitudinal latent class analysis. These features are planned
for a future version.

Overall, we intend the framework to bridge the different approaches to
clustering longitudinal data that exist from the various areas of
research. We encourage users and package developers to create interfaces
for their methods, as the availability of a standard framework for
performing a longitudinal cluster analysis lowers the barrier to
evaluating and comparing methods for applied researchers.

\hypertarget{sec:technical}{%
\section*{Computational details}\label{sec:technical}}
\addcontentsline{toc}{section}{Computational details}

The examples and figures in this paper were obtained using \texttt{R}
4.3.2 \citep{rcoreteam2021r} with the packages \texttt{latrend} 1.6.0,
\texttt{ggplot2} 3.4.4 \citep{Wickham2016ggplot2}, and
\texttt{data.table} 1.15.0 \citep{Dowle2020data.table}. The KmL method
was estimated based on the \texttt{kml} 2.4.6.1 package. The
distance-based method utilized the \texttt{dtwclust} 5.5.12 package. The
GBTM and GMM analyses were performed using the \texttt{lcmm} 2.1.0
package, with the parallel computation achieved using the
\texttt{foreach} 1.5.2 package \citep{weston2022foreach}.

\texttt{R} and all packages used within the article and the
\texttt{latrend} package are available from the Comprehensive \texttt{R}
Archive Network (CRAN) at (\url{https://CRAN.R-project.org}).

\hypertarget{acknowledgments}{%
\section*{Acknowledgments}\label{acknowledgments}}
\addcontentsline{toc}{section}{Acknowledgments}

This work was supported by Philips Research, Eindhoven, the Netherlands.
Niek Den Teuling and Steffen Pauws are employees of Philips. We are
grateful for the feedback provided by the anonymous reviewers. The
development of this framework builds upon the work of the \texttt{R}
community. The authors would like to express their appreciation for the
numerous longitudinal cluster packages that have been developed, as
these packages have made \texttt{R} a versatile platform for clustering
longitudinal data. Moreover, we gratefully incorporated many of the
cluster metrics by using the packages \texttt{clusterCrit}
\citep{Desgraupes2018clusterCrit} and \texttt{mclustcomp}
\citep{You2018mclustcomp}.

\bibliographystyle{apalike}
\bibliography{refs.bib}

\begin{thebibliography}{}

\bibitem[Adepeju et~al., 2020]{Adepeju2020akmedoids}
Adepeju, M., Langton, S., and Bannister, J. (2020).
\newblock Akmedoids {R} package for generating directionally-homogeneous
  clusters of longitudinal data sets.
\newblock {\em Journal of Open Source Software}, 5(56):2379.

\bibitem[Aghabozorgi et~al., 2015]{aghabozorgi2015time}
Aghabozorgi, S., Shirkhorshidi, A.~S., and Wah, T.~Y. (2015).
\newblock Time-series clustering - {A} decade review.
\newblock {\em Information Systems}, 53:16--38.

\bibitem[Arbelaitz et~al., 2013]{arbelaitz2013extensive}
Arbelaitz, O., Gurrutxaga, I., Muguerza, J., P{\'e}rez, J.~M., and Perona, I.
  (2013).
\newblock An extensive comparative study of cluster validity indices.
\newblock {\em Pattern recognition}, 46(1):243--256.

\bibitem[Babbin et~al., 2015]{babbin2015identifying}
Babbin, S.~F., Velicer, W.~F., Aloia, M.~S., and Kushida, C.~A. (2015).
\newblock Identifying longitudinal patterns for individuals and subgroups: An
  example with adherence to treatment for obstructive sleep apnea.
\newblock {\em Multivariate Behavioral Research}, 50(1):91--108.

\bibitem[Bates et~al., 2015]{Bates2015Fitting}
Bates, D., M{\"a}chler, M., Bolker, B., and Walker, S. (2015).
\newblock Fitting linear mixed-effects models using {lme4}.
\newblock {\em Journal of Statistical Software}, 67(1):1--48.

\bibitem[Benaglia et~al., 2009]{benaglia2009mixtools}
Benaglia, T., Chauveau, D., Hunter, D.~R., and Young, D. (2009).
\newblock {mixtools}: An {R} package for analyzing finite mixture models.
\newblock {\em Journal of Statistical Software}, 32(6):1--29.

\bibitem[Bouveyron, 2015]{Bouveyron2015funFEM}
Bouveyron, C. (2015).
\newblock {\em {funFEM}: Clustering in the Discriminative Functional Subspace}.

\bibitem[Cayanan et~al., 2019]{cayanan2019review}
Cayanan, E.~A., Bartlett, D.~J., Chapman, J.~L., Hoyos, C.~M., Phillips, C.~L.,
  and Grunstein, R.~R. (2019).
\newblock A review of psychosocial factors and personality in the treatment of
  obstructive sleep apnoea.
\newblock {\em European Respiratory Review}, 28(152).

\bibitem[De~la Cruz-Mes{\'i}a et~al., 2008]{de2008model}
De~la Cruz-Mes{\'i}a, R., Quintana, F.~A., and Marshall, G. (2008).
\newblock Model-based clustering for longitudinal data.
\newblock {\em Computational Statistics \& Data Analysis}, 52(3):1441--1457.

\bibitem[Den~Teuling et~al., 2021a]{denteuling2021comparison}
Den~Teuling, N.~G., Pauws, S.~C., and Van~den Heuvel, E.~R. (2021a).
\newblock A comparison of methods for clustering longitudinal data with slowly
  changing trends.
\newblock {\em Communications in Statistics - Simulation and Computation}.

\bibitem[Den~Teuling et~al., 2021b]{denteuling2021latent}
Den~Teuling, N.~G., Van~den Heuvel, E.~R., Aloia, M.~S., and Pauws, S.~C.
  (2021b).
\newblock A latent-class heteroskedastic hurdle trajectory model: Patterns of
  adherence in obstructive sleep apnea patients on {CPAP} therapy.
\newblock {\em BMC Medical Research Methodology}, 21(1):1--15.

\bibitem[Desgraupes, 2018]{Desgraupes2018clusterCrit}
Desgraupes, B. (2018).
\newblock {\em {clusterCrit}: Clustering Indices}.

\bibitem[Dowle and Srinivasan, 2020]{Dowle2020data.table}
Dowle, M. and Srinivasan, A. (2020).
\newblock {\em {data.table}: Extension of {`{data.frame}'}}.

\bibitem[Dziak et~al., 2015]{dziak2015modeling}
Dziak, J.~J., Li, R., Tan, X., Shiffman, S., and Shiyko, M.~P. (2015).
\newblock Modeling intensive longitudinal data with mixtures of nonparametric
  trajectories and time-varying effects.
\newblock {\em Psychological Methods}, 20(4):444--469.

\bibitem[Genolini et~al., 2015]{genolini2015kml}
Genolini, C., Alacoque, X., Sentenac, M., and Arnaud, C. (2015).
\newblock {kml} and {kml3d}: {R} packages to cluster longitudinal data.
\newblock {\em Journal of Statistical Software}, 65(4):1--34.

\bibitem[Gr{\"u}n and Leisch, 2008]{gruen2008flexMix}
Gr{\"u}n, B. and Leisch, F. (2008).
\newblock {FlexMix} version 2: Finite mixtures with concomitant variables and
  varying and constant parameters.
\newblock {\em Journal of Statistical Software}, 28(4):1--35.

\bibitem[Hamaker, 2012]{hamaker2012researchers}
Hamaker, E.~L. (2012).
\newblock Why researchers should think "within-person": {A} paradigmatic
  rationale.
\newblock In Mehl, M.~R. and Conner, T.~S., editors, {\em Handbook of Research
  Methods for Studying Daily Life}, pages 43--61. Guilford Publications.

\bibitem[Hennig, 2007]{hennig2007clusterwise}
Hennig, C. (2007).
\newblock Cluster-wise assessment of cluster stability.
\newblock {\em Computational Statistics \& Data Analysis}, 52(1):258--271.

\bibitem[Hubert and Arabie, 1985]{hubert1985comparing}
Hubert, L. and Arabie, P. (1985).
\newblock Comparing partitions.
\newblock {\em Journal of Classification}, 2(1):193--218.

\bibitem[Kom{\'a}rek, 2009]{Komarek2009New}
Kom{\'a}rek, A. (2009).
\newblock A new {R} package for {B}ayesian estimation of multivariate normal
  mixtures allowing for selection of the number of components and
  interval-censored data.
\newblock {\em Computational Statistics \& Data Analysis}, 53(12):3932--3947.

\bibitem[Liao, 2005]{liao2005clustering}
Liao, T.~W. (2005).
\newblock Clustering of time series data---a survey.
\newblock {\em Pattern Recognition}, 38(11):1857--1874.

\bibitem[McNicholas and Murphy, 2010]{mcnicholas2010model}
McNicholas, P.~D. and Murphy, T.~B. (2010).
\newblock Model-based clustering of longitudinal data.
\newblock {\em Canadian Journal of Statistics}, 38(1):153--168.

\bibitem[{Microsoft} and Weston, 2022]{weston2022foreach}
{Microsoft} and Weston, S. (2022).
\newblock {\em {foreach}: Provides Foreach Looping Construct}.

\bibitem[Muth{\'e}n, 2004]{muthen2004latent}
Muth{\'e}n, B. (2004).
\newblock Latent variable analysis: {G}rowth mixture modeling and related
  techniques for longitudinal data.
\newblock In {\em The {SAGE} Handbook of Quantitative Methodology for the
  Social Sciences}, pages 346--369. {SAGE} Publications, Inc.

\bibitem[Nagin, 2005]{nagin2005group}
Nagin, D.~S. (2005).
\newblock {\em Group-Based Modeling of Development}.
\newblock Harvard University Press, 1st edition.

\bibitem[Nagin et~al., 2018]{nagin2018group}
Nagin, D.~S., Jones, B.~L., Passos, V.~L., and Tremblay, R.~E. (2018).
\newblock Group-based multi-trajectory modeling.
\newblock {\em Statistical Methods in Medical Research}, 27(7):2015--2023.

\bibitem[Nielsen, 2018]{Nielsen2018crimCV}
Nielsen, J.~D. (2018).
\newblock {\em {crimCV}: Group-Based Modelling of Longitudinal Data}.

\bibitem[Proust-Lima et~al., 2017]{proustlima2017estimation}
Proust-Lima, C., Philipps, V., and Liquet, B. (2017).
\newblock Estimation of extended mixed models using latent classes and latent
  processes: The {R} package {lcmm}.
\newblock {\em Journal of Statistical Software}, 78(2):1--56.

\bibitem[{R Core Team}, 2022]{rcoreteam2021r}
{R Core Team} (2022).
\newblock {\em {R}: {A} Language and Environment for Statistical Computing}.
\newblock {R} Foundation for Statistical Computing.

\bibitem[Rousseeuw, 1987]{rousseeuw1987silhouettes}
Rousseeuw, P.~J. (1987).
\newblock Silhouettes: A graphical aid to the interpretation and validation of
  cluster analysis.
\newblock {\em Journal of Computational and Applied Mathematics}, 20:53--65.

\bibitem[Sard{\'a}-Espinosa, 2019]{sardaespinosa2019time}
Sard{\'a}-Espinosa, A. (2019).
\newblock Time-series clustering in {R} using the {dtwclust} package.
\newblock {\em The R Journal}.

\bibitem[Scrucca et~al., 2016]{Scrucca2016mclust}
Scrucca, L., Fop, M., Murphy, T.~B., and Raftery, A.~E. (2016).
\newblock {mclust} 5: Clustering, classification and density estimation using
  {G}aussian finite mixture models.
\newblock {\em The {R} Journal}, 8(1):205--233.

\bibitem[Van~de Schoot et~al., 2017]{van2017grolts}
Van~de Schoot, R., Sijbrandij, M., Winter, S.~D., Depaoli, S., and Vermunt,
  J.~K. (2017).
\newblock The {GRoLTS}-checklist: Guidelines for reporting on latent trajectory
  studies.
\newblock {\em Structural Equation Modeling: A Multidisciplinary Journal},
  24(3):451--467.

\bibitem[Van~der Nest et~al., 2020]{vandernest2020overview}
Van~der Nest, G., Lima~Passos, V., Candel, M.~J., and Van~Breukelen, G.~J.
  (2020).
\newblock An overview of mixture modelling for latent evolutions in
  longitudinal data: {M}odelling approaches, fit statistics and software.
\newblock {\em Advances in Life Course Research}, 43:100323.

\bibitem[Van~Dongen, 2000]{dongen2000performance}
Van~Dongen, S. (2000).
\newblock Performance criteria for graph clustering and {M}arkov cluster
  experiments.
\newblock techreport INS-R0012, CWI (Centre for Mathematics and Computer
  Science).

\bibitem[Venables and Ripley, 2002]{venables2002modern}
Venables, W.~N. and Ripley, B.~D. (2002).
\newblock {\em Modern Applied Statistics with {S}}.
\newblock Springer-Verlag, 4th edition.

\bibitem[Wickham, 2016]{Wickham2016ggplot2}
Wickham, H. (2016).
\newblock {\em {ggplot2}: Elegant Graphics for Data Analysis}.
\newblock Springer-Verlag New York, 2nd edition.

\bibitem[Yi et~al., 2022]{yi2022identifying}
Yi, H., Dong, X., Shang, S., Zhang, C., Xu, L., and Han, F. (2022).
\newblock Identifying longitudinal patterns of {CPAP} treatment in {OSA} using
  growth mixture modeling: Disease characteristics and psychological
  determinants.
\newblock {\em Frontiers in Neurology}, 13:1063461.

\bibitem[You, 2018]{You2018mclustcomp}
You, K. (2018).
\newblock {\em {mclustcomp}: Measures for Comparing Clusters}.

\end{thebibliography}

\end{document}